\documentclass[]{article}
\usepackage{a4wide}
\usepackage{blindtext}
\usepackage{todonotes}
\usepackage{listings}
\usepackage{color,soul}
\usepackage{graphicx}
\usepackage{subcaption}
\usepackage{float}
\usepackage{enumitem}   
\usepackage{bm}
\usepackage[norelsize]{algorithm2e}
\usepackage{cleveref}
\usepackage[bottom=0.8in,top=0.8in]{geometry}

\usepackage[utf8]{inputenc}
\usepackage[sorting=none]{biblatex}
\def\toptitlebar{\hrule height4pt\vskip .25in\vskip-\parskip}
\def\bottomtitlebar{\vskip .29in\vskip-\parskip\hrule height1pt\vskip
.09in} %

\begin{document}

\setcounter{secnumdepth}{4}
\setcounter{tocdepth}{4}
\pagenumbering{roman}

\title{\vspace{0.0cm}{
    \toptitlebar
    \textbf{sketch2code: Generating a website from a paper mockup}}
    \bottomtitlebar
}

\author{\textbf{Alexander Robinson}\\
    University of Bristol\\
    \texttt{ar15247@bristol.ac.uk}
}

\date{May 2018}

\maketitle


\begin{figure}[H]
    \centering
    \begin{subfigure}[H]{0.3\textwidth}
        \includegraphics[width=\textwidth]{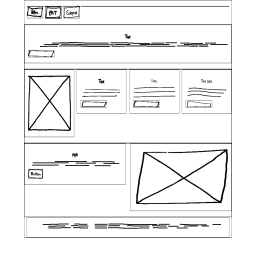}
        \caption{Original drawing}
        \label{figure:approach2_od}
    \end{subfigure}
    \begin{subfigure}[H]{0.3\textwidth}
        \includegraphics[width=\textwidth]{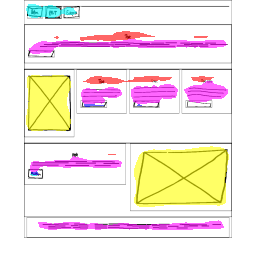}
        \caption{Segmented version}
        \label{figure:approach2_seg}
    \end{subfigure}
    \begin{subfigure}[H]{0.3\textwidth}
        \includegraphics[width=\textwidth]{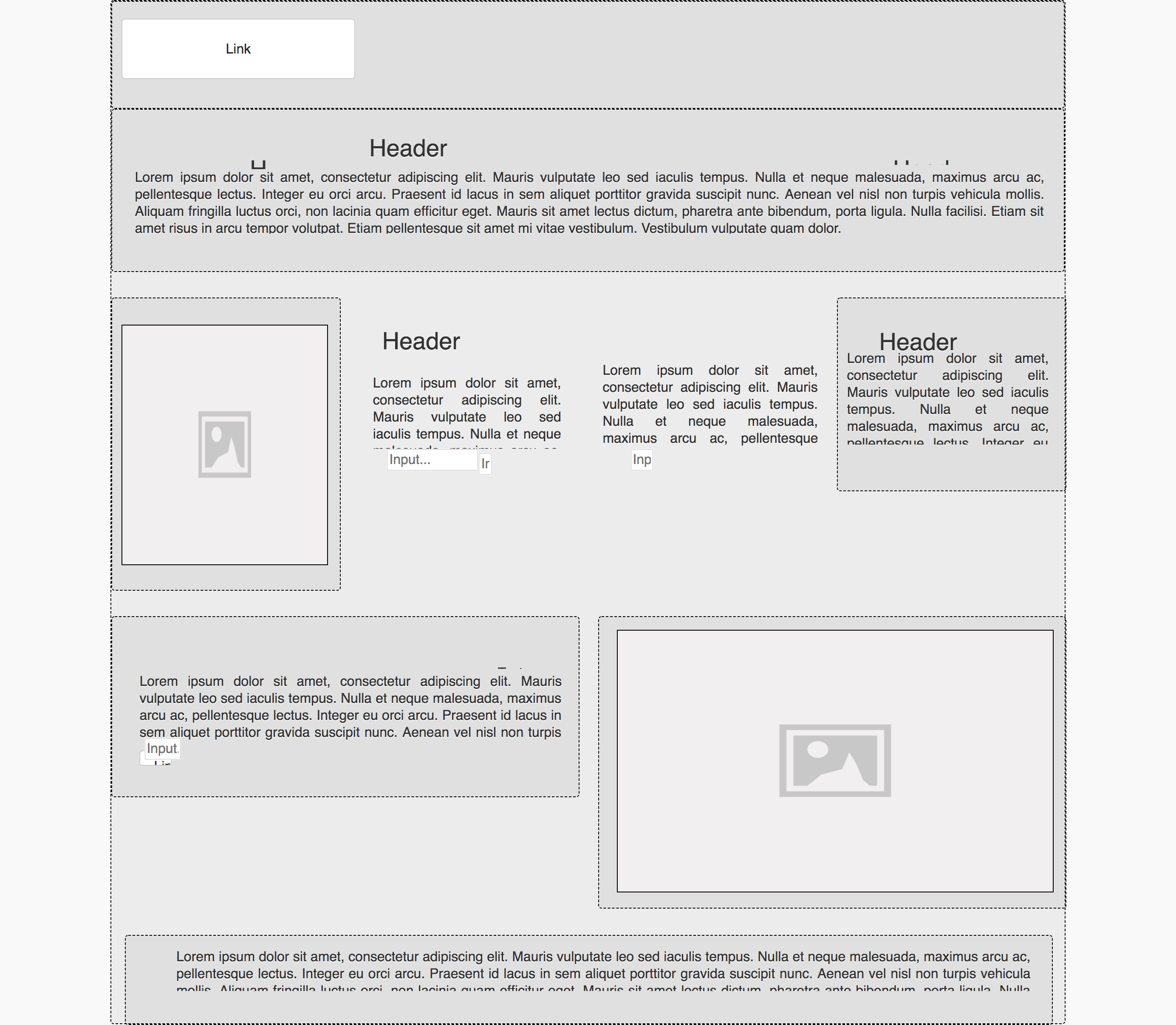}
        \caption{Rendered HTML}
        \label{figure:approach2_html}
    \end{subfigure}
\end{figure}

\vfill

\begin{center}
A dissertation presented for the degree of Bachelor of Science

\begin{figure}[H]
    \centering
    \includegraphics[width=0.4\textwidth]{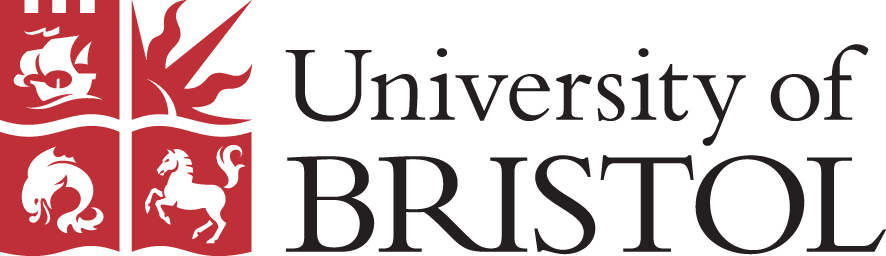}
\end{figure}

Department of Computer Science\\
May 2018
\end{center}

\clearpage
\section*{Abstract}
An early stage of developing user-facing applications is creating a wireframe to layout the interface \cite{Practitioner_Tools_and_Workstyles, designers_still_use_paper}. Once a wireframe has been created it is given to a developer to implement in code. Developing boiler plate user interface code is time consuming work but still requires an experienced developer \cite{User_Centered_Design_and_Agile_Methods}. In this dissertation we present two approaches which automates this process, one using classical computer vision techniques, and another using a novel application of deep semantic segmentation networks. We release a dataset of websites which can be used to train and evaluate these approaches. Further, we have designed a novel evaluation framework which allows empirical evaluation by creating synthetic sketches. Our evaluation illustrates that our deep learning approach outperforms our classical computer vision approach and we conclude that deep learning is the most promising direction for future research.

\clearpage
\section*{Declaration}
This dissertation is submitted to the University of Bristol in accordance with the requirements of the degree of Bachelor of Science in the Faculty of Engineering. It has not been submitted for any other degree or diploma of any examining body. Except where specifically acknowledged, it is all the work of the Author. \\ \\
Alexander Robinson, May 2018.

\clearpage
\section*{Acknowledgements}
I would like to thank Dr. Tilo Burghardt for all of his  invaluable guidance and support throughout the project.

\clearpage
\tableofcontents
\clearpage

\pagenumbering{arabic}

\part{Introduction}

An early step in creating an application is to sketch a wireframe on paper blocking out the structure of the interface \cite{Practitioner_Tools_and_Workstyles, designers_still_use_paper}. Designers face a challenge when converting their wireframe into code, this often involves passing the design to a developer and having the developer implement the boiler plate \textit{graphical user interface} (GUI) code. This work is time consuming for the developer and therefore costly \cite{User_Centered_Design_and_Agile_Methods}. 

Problems in the domain of turning a design into code have been tackled before: SILK \cite{SILK} turns digital drawings into application code using gestures; DENIM \cite{DENIM} augments drawings to add interaction; REMAUI \cite{REMAUI} converts high fidelity screenshots into mobile apps. Many of these applications rely on classical computer vision techniques to perform detection and classification.

We have identified a gap in the research which attempts to solve the over archiving problem. An application which translates wireframe sketches directly into code. This application considerable benefits:

\begin{itemize}
    \item Faster iteration - a wireframe can move to a website prototype with only the designers involvement.
    \item Accessibility - allows non developers to create applications.
    \item Removes requirement on developer for initial prototypes, allowing developers to focus on the application logic rather then boiler plate GUI code.
\end{itemize}

Furthermore, we have identified that deep learning methods may be applicable to this task. Deep learning has shown considerable success over classical techniques when applied to other domains, particularly in vision problems \cite{DBLP:journals/corr/DongLHT15,DBLP:journals/corr/VaradarajanTVN15,DBLP:journals/corr/VinyalsTBE14,DBLP:journals/corr/KarpathyF14,DBLP:journals/corr/GatysEB15a}. We hypothesis that a novel application of deep learning methods to this task may increase performance over classical computer vision techniques.

As such, the goal of this dissertation is two fold: a) create an application which translates a wireframe directly into code; b) compare classical computer vision techniques with deep learning methods in order to maximize performance.

This task involves major challenges:

\begin{itemize}
    \item Building both a deep learning and classical computer vision approach which can:
        \begin{itemize}
        \item Detect and classify wireframe elements sketched on paper
        \item Adjust the layout to fix for human errors from sketching
        \item Translate detected elements into application code
        \item Display the result to the user in an easy to use manor
        \end{itemize}
    \item Building a dataset of wireframes and application code
    \item Empirically evaluating the performance
\end{itemize}

This work is significant as we are address two research gaps: a) Researching methods to translate a wireframe into code and b) a novel application of deep learning to this domain.



In section \ref{section:background} we describe the background to this problem. We detail specific techniques we employee and the motivation behind using these techniques for this problem. In section \ref{section:method} we describe our dataset we created and utilised in both approaches and evaluation and we also explain our framework and two approaches. Finally, we describe our evaluation method and results, and conclude in sections \ref{section:evaluation} and \ref{section:conclusion_part}.

\clearpage
\part{Background} \label{section:background}

In this section we describe how the design process works and why an application which translates sketches into code is useful. We then explain why we have chosen to focus on websites for this dissertation, as well as explaining the challenges websites create. We move on to describe classical computer vision techniques and then machine learning techniques which we use in our method. Finally, we layout the research context of our work by describing related work and why this research is significant.

\section{The design process}  

\subsection{How applications are designed}

\begin{figure}[H]
    \centering
     \includegraphics[width=\textwidth]{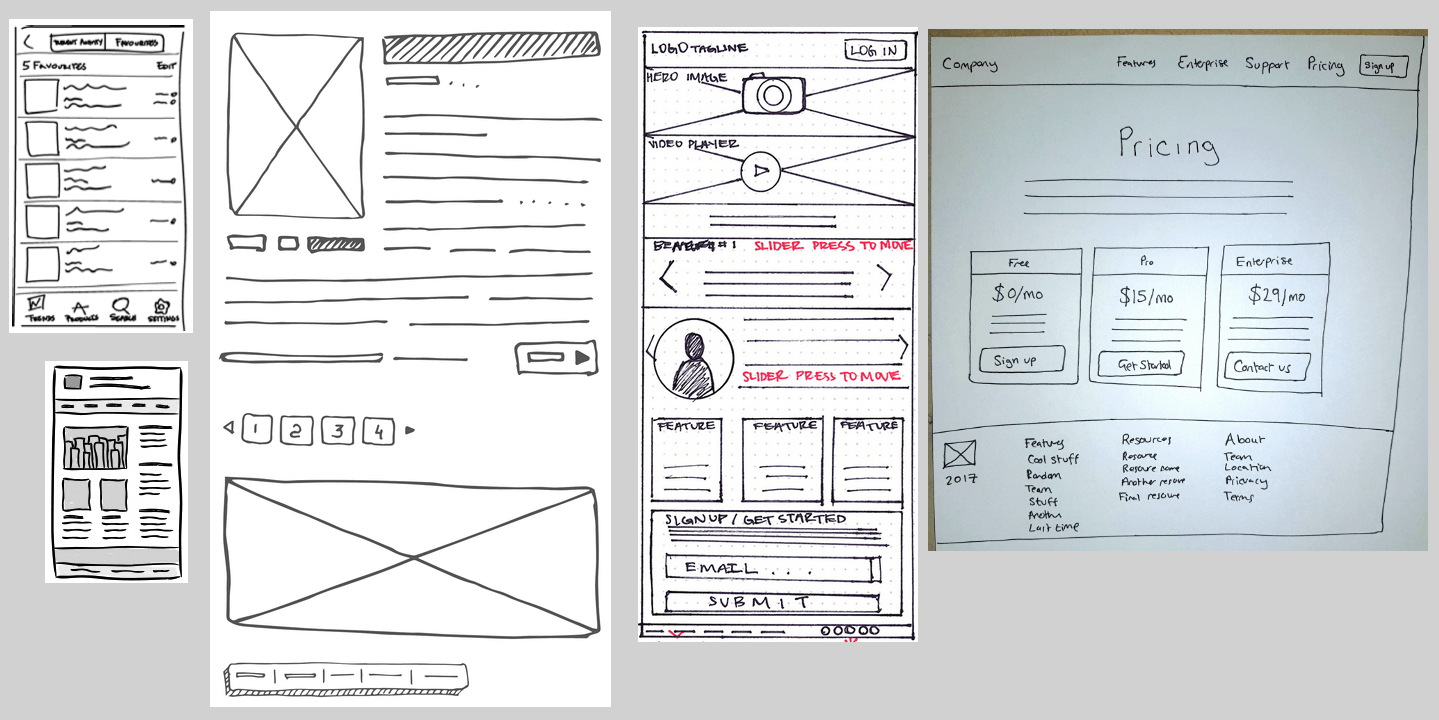}
    \caption{Examples of sketched wireframes for mobile and desktop applications. Notice that there are slight differences in styles but there are common symbols such as using horizontal lines to represent text.}
    \label{fig:example_wireframe}
\end{figure}

While the design process varies from individual to individual, for many projects it often starts as a digital or sketched wireframe \cite{Practitioner_Tools_and_Workstyles}. A wireframe is a document which outlines the basic structure of the application. A wireframe is a \textit{low fidelity} design document as it does not define specific details such as colours. After a wireframe is created it is reviewed and more detail is added i.e. it becomes a \textit{higher fidelity} mockup \cite{User_Centered_Design_and_Agile_Methods}. After the design is finalised it is implemented by a developer.

This process is time-consuming and can involve multiple parties. If a designer wishes to create a website they must work out all the details before having it implemented by a developer, as it is considerably easier to try out ideas in the design before they are converted into code. Further, translating a design into code is time consuming and developers are expensive.

Our proposed application reduces the time and cost factor by directly translating a wireframe into application code. It may be argued that the lengthy design process is intended to focus discussion on the overall structure before details. However, tools such as Balsamiq \cite{Balsamiq} or Wirify \cite{Wirify} are widely used and add filters to digital mockups to reduce the details thus showing that this is not an issue. On top of saving time and cost, the benefits of a generated website include:

\begin{itemize}
    \item Easier collaboration - a website can be instantly hosted and shared for others to review
    \item Interactivity - unlike digital images, a website can add interactivity such as buttons and forms
    \item No middle people - developers often have to interpret aspects missing from a design, by allowing a designer to directly implement the website the designer can add these details.
\end{itemize}

\subsection{Wireframes} \label{section:background_sketches}

\begin{figure}[H]
    \centering
    \begin{subfigure}[H]{0.3\textwidth}
        \includegraphics[width=\textwidth]{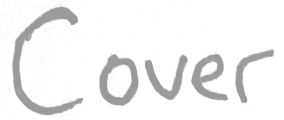}
        \caption{A title element: text with no container box}
    \end{subfigure}
    \begin{subfigure}[H]{0.3\textwidth}
        \includegraphics[width=\textwidth]{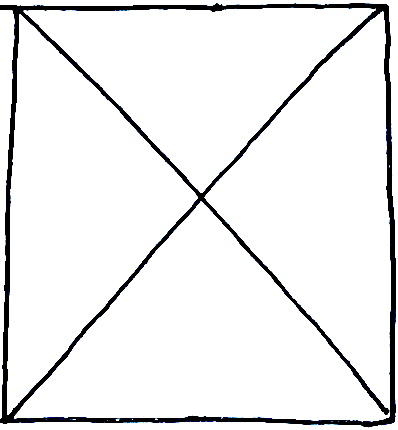}
        \caption{An image element: a box with a cross through it}
    \end{subfigure}
    \begin{subfigure}[H]{0.3\textwidth}
        \includegraphics[width=\textwidth]{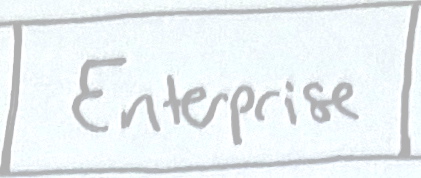}
        \caption{A button element: text with a tight container box}
    \end{subfigure}
    \begin{subfigure}[H]{0.3\textwidth}
        \includegraphics[width=\textwidth]{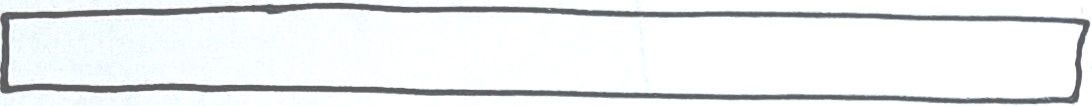}
        \caption{An input element: an empty container box with a small aspect ratio}
    \end{subfigure}
    \begin{subfigure}[H]{0.3\textwidth}
        \includegraphics[width=\textwidth]{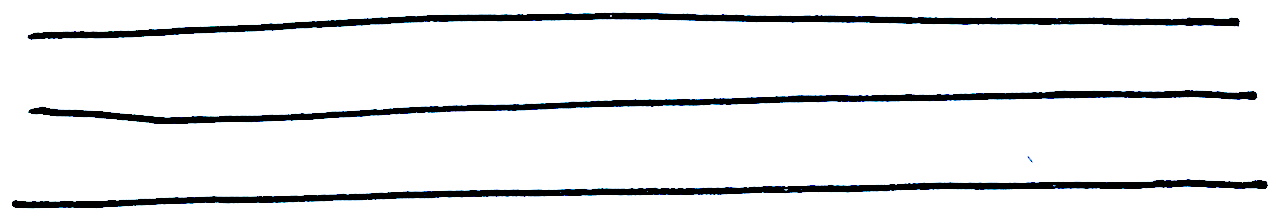}
        \caption{A paragraph element: three or more lines of the same length positioned on top of one another}
    \end{subfigure}
    \caption{Examples from each of the five elements we use to represent title, image, button, input, and paragraph elements in our wireframes. These symbols are based off popular and commonly understood wireframe elements.}\label{fig:wireframe_elements}
\end{figure}

Although there is no agreed standard, wireframe sketches often use a similar set of symbols which have commonly understood meanings. For the purpose of this dissertation we will focus on five symbols: images, titles, paragraphs, buttons, and input elements (figure \ref{fig:wireframe_elements}). 

Despite the existence of special purpose digital wireframe tools \cite{Balsamiq,mockup_tools,,photoshop,InVision,UXPin,Marvel} many designers still start on paper \cite{Practitioner_Tools_and_Workstyles,designers_still_use_paper}, sketching with pencil, and then later digitize \cite{Rough_and_Ready_Prototypes}. According to Myers this is primarily explained by the designers background being in art, the designer feeling restricted by digital tools, or simply that it is easiest, with no need to learn any software \cite{designers_still_use_paper}. As such, we intend to focus our application on translating wireframe sketches drawn on paper into application code.

\section{Website development}

In this section we explain why we focus on websites, we describe how websites are structured, and the challenges of dealing with websites. These are important to understand for our dataset processing steps (section \ref{section:dataset}) and for design choices in our framework (section \ref{section:framework}).

We focus on producing the code for the front end of websites rather than desktop or mobile applications, as:

\begin{itemize}
    \item All websites are built in the same language, \textit{hypertext markup language} (HTML) while mobile and desktop applications can be build in a number of languages.
    \item Websites remain one of the most popular tools for building applications.
    \item We require a large dataset for our method, building this dataset from websites is easier than mobile or desktop applications as they are more easily accessible.
    \item Effectively evaluating mobile or desktop applications by running them requires significant resources as it requires the emulation of devices, while websites can be run more easily.
\end{itemize}

\begin{figure}[H]
    \centering
    \begin{subfigure}[H]{\textwidth}
        \begin{lstlisting}
<!DOCTYPE html>
<html>
  <head>
    <style>
      body {
        background-color: rgb(254, 196, 3);
      }
      .container {
        text-align: center;
      }
    </style>
  </head>
  <body>
    <div class='container'>
      <h1>My First Heading</h1>
      <p>My first paragraph.</p>
      <img src="dog.jpg" height="142">
    </div>
  </body>
</html>
        \end{lstlisting}
         \caption{HTML source code}
    \end{subfigure}
    \begin{subfigure}[H]{0.5\textwidth}
        \includegraphics[width=\textwidth]{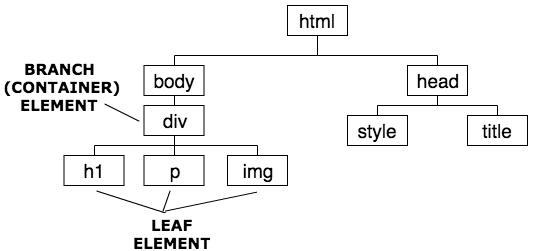}
         \caption{Document object model tree}
    \end{subfigure}
    \begin{subfigure}[H]{0.35\textwidth}
        \includegraphics[width=\textwidth]{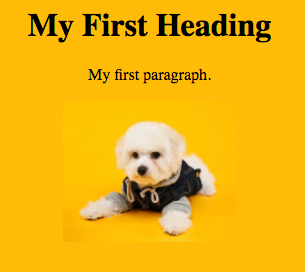}
        \caption{Rendered webpage}
    \end{subfigure}
    
    \caption{An example website with HTML and CSS showing three levels of abstraction (a), (b), and (c). (a) shows how HTML consists of nested elements which describe the structure of elements. b) shows how the HTML represents a GUI tree with leaf elements which contain content and branch (container) elements which group elements. (c) shows how the HTML is interpreted by a rendering engine to produce a visual result.}
    \label{fig:example_webpage}
\end{figure}

Websites are composed structurally with HTML element tags and styled with \textit{Cascading Style Sheets} (CSS). The structure of a website consists of the types of elements e.g. table, paragraph, button, and how they are positioned relative to one another. The style consists of the colours, fonts, and borders. HTML is constructed as a tree of objects known as the \textit{document object model} (DOM). Branches of this tree represent containers, examples of these are \lstinline{<div>}, \lstinline{<footer>}, \lstinline{<header>}, leaves of the tree are elements which contain content e.g. \lstinline{<img>}, \lstinline{<p>}, \lstinline{<button>}.

HTML rendering engines do not strictly enforce HTML standards, and this leads to poor and sometimes invalid code. As such, in order to understand the necessity of our dataset processing method (section \ref{section:dataset}), it is important to understand aspects of HTML which lead to poor code:

\begin{itemize}
    \item Unlabeled elements: HTML consists of semantic and functional elements. Most elements are functional and instruct the rendering engine to render in a specific way. However, a small set of elements are semantic and have no effect on the rendering of the webpage. Semantic elements such as \lstinline{<header>} and \lstinline{<footer>} are used by search engines to better understand a webpage. Because semantic elements do not change the rendering of a webpage they are often ignored but it is best practice to include them.
    \item HTML contains style and CSS contains structure: HTML can be used to style parts of a website (e.g. \lstinline{<strong>} tags) and CSS can be used to structure websites (e.g. the \lstinline{position} attribute). As such, extracting the structure from a webpage is not as simple as viewing only the HTML.
    \item JavaScript modifies webpage structure: JavaScript is a website scripting language and is run on a website after a website is rendered and can manipulate the DOM.
    \item Multiple structures lead to same rendering: there are many potential combinations of HTML which lead to the exact same rendered result. This highlights the difficulty in designing evaluation methods which compare the website structure (section \ref{section:evaluation}).
\end{itemize}

\section{Related Work} \label{section:related_work}

In this section we briefly examine the top research in design to code applications. We identify common problems which arise in this research. We then identify two gaps: little research utilising deep learning methods, and no sketch to code solution. Finally we explain why these gaps might exist.

The study of turning design documents into code is not new and until the recent introduction of machine learning techniques there has been little progression or adoption of these tools. There are two main research areas we will look at, turning low fidelity drawings (digital or analogue) into code, and turning very high fidelity wireframes/screenshots into code. These share a common goal - translating a design into an application - and many techniques. The major difference is the fidelity of the designs being translated, which leads to different challenges. Note that we limit our discussion to the areas stated above, and will not go into detail about the many related fields such as wireframe augmentation e.g. DENIM \cite{DENIM} or SketchWizard \cite{SketchWizard}.

We firstly examine existing research in low fidelity design to code applications. SILK \cite{SILK} and MobiDev \cite{MobiDev} are applications which focus on desktop and mobile respectively, both use a combination of computer vision techniques in order to classify shapes drawn on a digital surface into predefined components. Both applications have shown that using computer vision techniques such as contour detection, corner detection, and line detection you can robustly classify shapes as application elements such as buttons and text. We have identified problems with these approaches:

\begin{itemize}
    \item Using these techniques the applications are time consuming to extend with new elements as they rely on judgement of the programmer to engineer features to classify all drawn representations a user may input.
    \item Little work has been done in post-processing of detected components to correct for mistakes in the drawing (e.g. a user may not draw two items perfectly aligned when they meant to), this results in a crude representation of the sketch were elements are placed exactly where they are drawn rather then the program estimating the correct placement.
\end{itemize}

We now look at the related research in turning high fidelity designs to code, REMAUI \cite{REMAUI} is a leading example. These applications take high fidelity designs such as a wireframe image in photoshop or even a screenshot of an existing application and translate into code. These applications not only translate the structure, but also the style (images, colours, fonts) of the designs. REMAUI uses similar techniques to MobiDev to identify and classify elements from the image into a series of classes: images, text, title, buttons, menus, inputs. REMAUI also faces the problem that adding new examples of classes is challenging and time consuming. The problem is made worse by the inclusion of style in the designs which leads to a considerably higher variation compared to the digital drawings in the low fidelity designs.

Another application which turns high fidelity designs to code is pix2code \cite{pix2code}, it takes a screenshot and employs deep learning to predict the code. It was trained on a large synthetic corpus of graphical user interface (GUI) screenshots and associated code. It utilises a recurrent neural network to predict token by token the result. It learns the relationship between components and their code allowing it to generalize to unseen examples. Pix2code provides a number of benefits over previous techniques mainly that it only requires example data and can learn the association for new styles. However, despite pix2code's high performance on its synthetic dataset it is not capable of generalising to real world examples \cite{Moran2018MachineLP}.

We have identified a gap in the research: transforming wireframes sketched on paper directly into code. As far as we are aware this has not been attempted despite it providing a number of significant benefits by leapfrogging the lengthy wireframing and coding procedure which requires experienced developers and designers, as well as allowing anyone who can draw a wireframe on paper to create a fully functioning website with no training.

Further, deep learning has shown considerable success in vision problems \cite{DBLP:journals/corr/DongLHT15,DBLP:journals/corr/VaradarajanTVN15,DBLP:journals/corr/VinyalsTBE14,DBLP:journals/corr/KarpathyF14,DBLP:journals/corr/GatysEB15a}. We have identified a gap in research of applying deep learning techniques to low fidelity design to code applications. We believe deep learning is suited to this problem as this task is primarily a vision problem and there is lots of data.

\section{Computer vision techniques}

We use classical computer vision techniques to detect and classify wireframe elements from an image for our first approach (approach 1). This is inspired by existing research \cite{SILK,MobiDev,REMAUI,Moran2018MachineLP} which we analysed in the previous section. In this section we describe techniques to detect and classify elements, and justify their use over alternatives.

\subsection{Image denoising}

Our method uses real world images (as opposed to digitally created images) as input i.e. images taken from a web camera. These images often contain Gaussian noise due to variations of current over the sensor of a camera \cite{DBLP:journals/corr/BoyatJ15}. Noisy images can lead to worse performance in edge detection \cite{noisy} which we utilise for element detection. As such, it is important to reduce this noise.

Deep learning techniques such as denoising auto-encoders are very effective against noise removal but they are slower then kernel filtering techniques \cite{xie2012image}.

A Gaussian blur is a filtering technique which uses a Gaussian kernel to smooth an image, Gaussian blurs are highly effective against Gaussian noise \cite{shukla2014gaussian} but they are not edge preserving \cite{median_blur}.

A median blur is a filtering technique which uses a kernel which preserves the median value in each window. Median blurs are effective against Gaussian noise \cite{shukla2014gaussian} and preserve edges \cite{median_blur}.

One of our aims is to develop an application which can be used in real time, we therefore require fast denoising algorithms. Edges are an important feature of wireframe symbols and are used as features to detect and classify elements. As such, preserving edges is an important property of our denoising method. We therefore proceed by using median blurs for denoising as they are fast, effective, and edge preserving.

\subsection{Colour detection}

Colour detection is used in our method to aid element detection. In this section we explain different techniques of detecting colours in images. We focus on detection via thresholds as this is an efficient technique for large uniform colour blobs \cite{kadouf2013colour}.

Digital images are often represented with three colour channels, \textit{red, green, and blue} (RGB) known as a \textit{colour space}. Other colour spaces exists e.g. \textit{hue, saturation, and value} (HSV) \cite{HSV} or L*A*B \cite{HSV_VS_LAB}, which represent colour in different ways.

Threshold detection uses value thresholds for each channel to filter colours. For example, to detect only red objects, the green and blue channels can be zeroed on an RGB image. Colour spaces have properties which can aid colour detection, for example the HSV colour space is more robust towards external lighting changes compared with other colour spaces \cite{park2004pattern}. Our aim is to use real images and therefore invariance towards lighting changes is an important property which we make use of in approach 1 (section \ref{section:approach1}).

\subsection{Edge detection}

We are concerned with detecting elements from wireframe sketches. The wireframe element symbols (section \ref{section:background_sketches}) primarily consist of straight edges. Edge detection is an important technique we use in our detection of elements. In this section we explain our choice in edge detection method. There are many edge detection techniques, some common choices are:

\begin{itemize}
    \item The Sobel operator \cite{sobel} which computes the gradient map of an image, and then yields edges by thresholding the gradient map.
    \item The Canny operator \cite{canny1987computational} is an extension of the Sobel operators which adds Gaussian smoothing as a preprocessing step and uses the bi-threshold to get the edges.
    \item Richer Convolutional Features \cite{DBLP:journals/corr/LiuCHWB16} is a state of the art deep learning approach using convolutional networks.
\end{itemize}

While with real world images deep learning approaches perform considerably better \cite{DBLP:journals/corr/LiuCHWB16}, Canny is efficient in \textit{plain} images such as a black pen sketch on white paper \cite{zhou2005automatic}. Canny also performs highest (lowest number of false edges) compared with the Sobel, Prewitt, Roberts, Laplacian, and Zero Crossing  \cite{DBLP:journals/corr/KatiyarA14, edge_detection_canny_best}. As such, we proceed by using the Canny operator for edge detection.

\subsection{Segmentation}

In order to classify wireframe elements they must first be detected. A wireframe sketch will contain many elements and therefore a method of detecting element boundaries is required. There are many potential segmentation algorithms:

\begin{itemize}
    \item Colour based - region growing, and histogram-based methods all use colours to segment objects.
    \item Structure based - contour detection uses edge information and boundary following to distinguish segments. However, these methods fail for disconnected non-closed boundaries.
    \item Trainable segmentation - learns to segment objects based on examples. These techniques often utalise deep learning methods. We discuss these further in section \ref{section:segmentation}.
\end{itemize}

Our first approach uses classical computer vision techniques so we will not consider trainable segmentation as a segmentation object for approach 1. Our boundaries consist of the lines which represent the element symbols. We therefore chose to use structural based techniques which give better colour invariance. 

Our method makes heavy use of contour detection. Contours are the curve joining all continuous points with the same colour or intensity along a boundary. We utilise a simple boundary following algorithm Suzuki85 \cite{Contour_detection} which is fed edges from our edge detection method.

Further, we will make use of contour approximation with the Douglas-Peucker algorithm \cite{douglas1973algorithms}. This allows for imperfect shapes to be corrected by approximating a contour over slight mistakes in the shape.

\subsection{Text detection}

Our method involves detecting text from sketches, we use \textit{stroke width transform} (SWT) \cite{stroke-width-transform} to detect text. SWT is fast, language independent, lightweight scene text detector, note that it does not recognise text only detect. SWT is particularly advantageous in our method due to its speed and language independence.


Part of our method utilises text detection. Note that we are only concerned with detection and not recognition. Further, we are only concerned with detection of handwritten text and therefore do not consider template matching techniques. There has been a significant of research into scene text detection \cite{zhu2016scene}. We considered multiple approaches:

\begin{itemize}
    \item Deep learning methods - such as CNN based methods \cite{jiang2017r2cnn, wang2012end, buvsta2017deep, DBLP:journals/corr/LiaoSBWL16, DBLP:journals/corr/HeZYL17} perform well in both natural and artificial scenes but require training data to tune performance for specific styles of text and are often language dependent.
    \item Feature based methods - such as Canny Text Detector \cite{7780757} or Stroke width transform \cite{stroke-width-transform} engineer features specific to text in order to detect text in images.
\end{itemize}

We opted for \textit{stroke width transform} (SWT) \cite{stroke-width-transform} which is language independent, lightweight text detector, describe fully in section \ref{section:background}.

\section{Machine learning techniques}

Machine learning is a powerful technique which gives computer systems the ability to learn from data without being explicitly programmed. We consider machine learning a perfect tool for some problems in this domain as there is considerable data and many classification and detection tasks. In this section we describe two key techniques which we make use of in our method - multi layer perceptron networks and semantic segmentation networks - as well why these techniques are particularly applicable to this problem.

\subsection{Artificial Neural Networks}

\textit{Artificial Neural Networks} (ANNs) are a simplified computational model modeled after biological neurons found in the brain. An artificial neuron is a fundamental computational unit which has a collection of input connections and output connections. A neuron computes the weighted average of its inputs, this is passed through a nonlinear function known as an activation function, the result of which is passed along to all output connections. A series of connected neurons is known as a neural network and can be used for computation.

A \textit{feedforward} ANN is a network where data moves one way i.e. there are no loops, this is opposed to a recurrent ANN which contains cycles \ref{nasrabadi2007pattern}.

\subsection{Multilayer perceptron networks}

\textit{Multilayer perceptron networks} (MLPs) are a class of feedforward ANN which can distinguish data that is not linearly separable. It is important to understand how these networks operator in order to understand our method which makes heavy use of MLPs.

MLPs consist of multiple neurons arranged in layers. All neurons from adjacent layers have connections between them, we say the layer is \textit{fully connected}. Each connection has an associated weight. There are three types of layers:
\begin{itemize}
    \item Input layer - the first layer in the network, no processing is done, simply passes information to the next layer.
    \item Hidden layers - zero or more layers which sit after the input layer. These perform computations and transfer information from the input layer to the output layer.
    \item Output layer - responsible for computation and transferring the result of these computations out of the network.
\end{itemize}

When an MLP is training, input and output data pairs are fed into the MLP and an algorithm (backpropagation \cite{lecun1988theoretical}) updates the weights between neurons trying to model the computational transformation which transforms the input data into the output data. Often the more difficult the transformation, the more data is required to learn it, and the more complicated the network. When an MLP reaches an acceptable level of performance it can be run by providing data to the input layer and having the MLP produce an output.

The \textit{width} of a layer is the number of neurons in that layer. The wider a layer the better it can memorise information \cite{lecun1988theoretical}. The \textit{depth} of a network is the number of hidden layers. Deeper networks can learn features at different levels of abstraction \cite{lecun1988theoretical}.

\textit{Generalisation} refers to the performance of the network when it is fed new data. \textit{Overfitting} is a problem networks have when they model the relationship between the input and output data during training too rigidly, it results in poor generalisation \cite{lecun1988theoretical}. More complicated (wider and deeper) networks are often more prone to overfitting \cite{lecun1988theoretical}.

There are a number of other machine learning techniques which can often be used instead of MLPs, for example \textit{support vector machines} (SVM) \cite{hearst1998support} or \textit{random forests} \cite{breiman2001random}. As we discuss in section \ref{section:approach1}, the main problem we face is a multi class classification problem. SVMs are applicable for binary classification problem, while multiple SVMs can be used in parallel to perform a multi class classification we found MLPs simpler to implement. Random forests are another technique which could have been used but have similar performance to MLPs.

\subsection{Deep learning}

Deep learning is the field of machine learning which uses deep neural networks consisting of many hidden layers. Deep learning has shown immense success in certain fields often far outperforming classical methods \cite{DBLP:journals/corr/DongLHT15,DBLP:journals/corr/VaradarajanTVN15,DBLP:journals/corr/VinyalsTBE14,DBLP:journals/corr/KarpathyF14,DBLP:journals/corr/GatysEB15a}. We believe the same success of deep learning can be applied to this domain because there is lots of data - a requirement of most deep learning methods - and because this is primarily a vision problem and deep learning has become the de facto standard for top performance in many vision problems \cite{lecun2015deep}.

\subsection{Convolutional neural networks}

A high level understanding of \textit{Convolutional neural networks} (CNNs) \cite{lecun2015deep} is important to understand semantic segmentation which is core to our method. Furthermore, CNNs are important to understand as they are an alternative approach we assessed.

A specific class of neural network which loosely models the connectivity pattern between neurons found in the visual cortex of animals. Neurons respond to signals in a local region in the receptive field. The receptive fields of different neurons cover the entire visual field and partially overlap. CNN are particularly useful in image based tasks but can be applied to other tasks such as natural language processing.

CNNs are particularly useful over MLPs for image processing tasks as MLPs suffer from the curse of dimensionality due to the full connectivity between neurons and therefore do not scale well for high resolution images. Further, MLPs don’t take into account the spatial structure of data, so pixels on opposite sides of an image would be treated the same way as pixels close together.

CNNs in the image domain use raw pixel data as input. Like an MLP they consist of an input, output, and multiple hidden layers. The hidden layers typically consists of:

\begin{itemize}
    \item Convolutional layers - which apply a convolution operation to the input passing the result to the next layer, each convolutional neuron processes data only for its receptive field.
    \item Pooling layers - which combine outputs of neuron clusters at one layer into a single neuron in the next layer. The pooling layer serves to progressively reduce the spatial size of the representation in order to reduce the number of parameters and amount of computation in the network, which helps control overfitting.
    \item Fully connected layers - Connect every neuron in one layer with every neuron in another. These are placed at the end of the network and are used to produce the final output based on all of the features the network has derived.
\end{itemize}

CNNs have three major distinguishing features:

\begin{itemize}
    \item 3D volumes of neurons - the layers of a CNN have neurons arranged in 3 dimensions: width, height and depth. The neurons inside of each layer are connected to only a small region of the layer before it (the receptive field).
    \item Local connectivity - CNNs exploit spatial locality by enforcing a local connectivity pattern between neurons of adjacent layers. This ensures that the learnt "filters" produce the strongest response to a spatially local input pattern. Stacking many layers results in non-linear filters that become increasingly global.
    \item Shared weights - each filter is replicated across the entire visual field. These replicated units share the same parameterization and form a feature map. This results in all the neurons in a convolutional layer respond to the same feature within their specific response field. This allows translation invariant features.
\end{itemize}

These properties allow CNNs to better generalise on vision problems. Weight sharing reduces the total number of free parameters learned which results in lowering the memory requirements while running.

\subsection{Semantic Segmentation} \label{section:segmentation}

\begin{figure}[H]
    \centering
    \begin{subfigure}[H]{0.45\textwidth}
        \includegraphics[width=\textwidth]{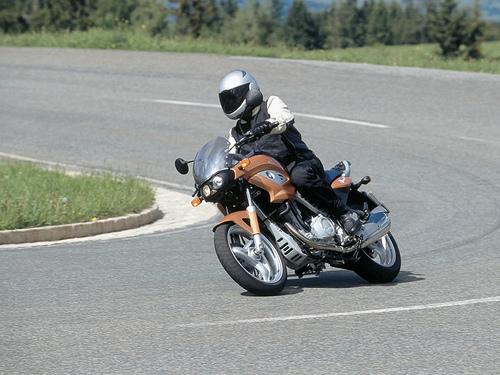}
        \caption{Input Image}
    \end{subfigure}
    \begin{subfigure}[H]{0.45\textwidth}
        \includegraphics[width=\textwidth]{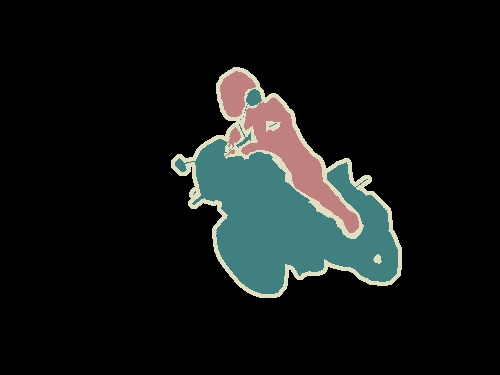}
        \caption{Segmented map}
    \end{subfigure}
    \caption{Semantic segmentation example. Source \cite{Everingham10}}
    \label{fig:segmentation_example}
\end{figure}

Image segmentation is a common image processing task, the goal is to simplify and/or change the representation of an image into something more meaningful. It involves labeling each pixel such that pixels which share characteristics are labeled together. There are a number of methods such as clustering methods, Texton Forests, or edge detection but we will focus on trainable segmentation networks as these techniques are best performing \cite{garcia2017review}.  Trainable Segmentation uses ANNs trained on a pair of data: the image, and a labeled segmented map of the image.

Automatic segmentation networks can not only segment objects but it can also classify and label segments. Automatic segmentation networks are based on CNNs, however pooling layers in CNNs discard the 'where' information which is required for precise object boundaries in segmentation. There are two main architectures segmentation networks are based onto get around this issue:

\begin{itemize}
    \item Encoder Decoder architecture
    \item Dilated / atrous convolutions
\end{itemize}

We will focus our discussion on dilated convolution architectures as architectures based on these are best performing \cite{deeplab}.

\begin{figure}[H]
    \centering
    \includegraphics[width=0.5\textwidth]{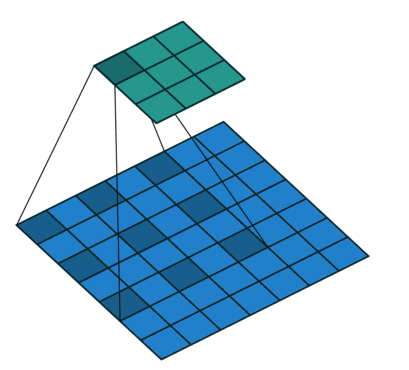}
    \caption{Dilated/Atrous Convolutions. Source \cite{dumoulin2016guide}}
    \label{fig:dilated_convolutions}
\end{figure}

A dilated convolution (also called a atrous convolution) \cite{DBLP:journals/corr/YuK15} is a convolution with defined gaps as shown in figure \ref{fig:dilated_convolutions}. Dilated convolutions support exponential expansion of the receptive field (without loss of resolution or coverage) while only growing the number of parameters logarithmically \cite{DBLP:journals/corr/ZhouWWZ15}. This means that it allows a larger receptive field with the same computation and memory costs as an encoder decoder architecture. A large receptive field is an important characteristic we require for our method as we require classification based on global image context (see section \ref{section:approach2}). Note that pooling and strided convolutions are similar concepts however, both reduce the resolution.

There are many segmentation networks \cite{DBLP:journals/corr/PengZYLS17,DBLP:journals/corr/BadrinarayananK15,DBLP:journals/corr/ZhaoSQWJ16} but we have chosen Deeplabv3+ (henceforth Deeplab) \cite{deeplab} due to its top performance \cite{leaderboard}. Deeplab is a state of the art network based on dilated convolutions. Deeplab v3+ (henceforth Deeplab) is the fourth iteration in the deeplab series \cite{DBLP:journals/corr/ChenPKMY14,DBLP:journals/corr/ChenPK0Y16,DBLP:journals/corr/ChenPSA17}. Another major advantage of Deeplab over other approaches is that it uses atrous spatial pyramid pooling. \textit{Atrous spatial pyramid pooling} (ASPP) allows robust segmentation of objects at multiple scales. ASPP probes an incoming convolutional feature layer with filters at multiple sampling rates and effective fields-of-views, thus capturing objects as well as image context at multiple scales.

\subsection{Fine tuning}

For many applications little training data is available. CNNs usually require a large amount of training data in order to avoid overfitting. A common technique is to train the network on a larger dataset from a related domain. Once the network parameters have converged an additional training step is performed using the in-domain data to fine tune the network weights. This allows CNNs to be successfully applied to problems with small training sets. 

\clearpage
\part{Method} \label{section:method}
The goal of this dissertation is twofold: a) create an application which translates a wireframe directly into code; b) compare classical computer vision techniques with deep learning methods in order to maximize performance.

We restricted our application to working on wireframes drawn with a dark marker on a white background. This is justified as wireframes are commonly designed on paper or whiteboards. We expected our application to produce and render code in real time. As explained in section \ref{section:background} we focus on websites. In order to achieve the goal of this dissertation we created two approaches:

\begin{itemize}
    \item Approach 1: using classical computer vision techniques
    \item Approach 2: using deep learning techniques 
\end{itemize}

In this section we first describe our dataset, and then our general framework which accepts an image of a wireframe, performs pre and post processing, and can use either approach to generate the website. We then explain approach in turn.

Note that specific parameters for operations (such as blurs, dilations, edge detection) can be found in our code. However, we highlight any significant values which require justification. 

\section{Dataset} \label{section:dataset}

Our approaches require a dataset which contains a wireframe sketch and associated website code. Sourcing a quality dataset is often a challenge in many machine learning projects. We were not aware of any dataset which contained wireframes sketches and associated website code, and therefore created our own. We considered three options to create such a dataset:

\begin{enumerate}[label=(\roman*)]
    \item Finding websites and manually sketching them. \label{item:dataset_method_1}
    \item Manually sketching websites and building the matching website. \label{item:dataset_method_2}
    \item Finding websites and automatically sketching them. \label{item:dataset_method_3}
\end{enumerate} 

Deep learning methods require a sufficiently large dataset with hundreds of samples. Even with data augmentation techniques (augment smaller dataset into larger by modifying current samples) \ref{item:dataset_method_1} and \ref{item:dataset_method_2} would require significant resources to complete. In addition, human error and different human opinions on how exactly to create the sketches could undermine the quality of the dataset. As result we use method \ref{item:dataset_method_3}.

Our aim is to work generally on any website wireframe and therefore we required a dataset which represented a wide variety of wireframes. As such, we used real websites rather then a synthetic approach like that used by pix2code \cite{pix2code_dataset}. Our process involved collecting pairs of a sketched version of the website as well as the structural code of the website. 

\begin{figure}[H] 
    \centering
    \begin{subfigure}[H]{1\textwidth}
        \centering
        \includegraphics[width=0.5\textwidth]{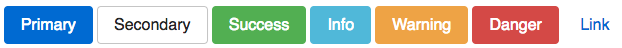}
    \end{subfigure}
    \begin{subfigure}[H]{1\textwidth}
        \centering
        \includegraphics[width=0.5\textwidth]{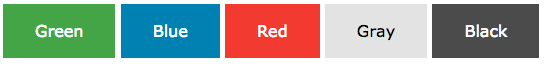}
    \end{subfigure}
    \begin{subfigure}[H]{1\textwidth}
        \centering
        \includegraphics[width=0.5\textwidth]{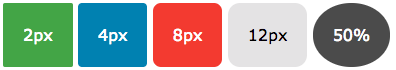}
    \end{subfigure}
    \caption{Web pages contain structure from HTML and style from CSS. We show a sample of \lstinline{<button>} elements with various styles. This highlights the need for a processing step to normalise web page style.}
    \label{fig:button_styles}
\end{figure} 

We identified a number of challenges to collecting the structure from websites:

\begin{enumerate}[label=(\alph*)]
    \item Wireframes have a small element set, HTML has a much larger element set, a simplified representation is required e.g. HTML has both an \lstinline{<img>} and \lstinline{<svg>} tag but both represent a graphical element. \label{item:dataset_challenge1}
    \item Wireframes are style invariant, and therefore we require only the structure. Websites have many different styles for elements e.g. a \lstinline{<button>} element (figure \ref{fig:button_styles}). \label{item:dataset_challenge2}
    \item Wireframes are static however, website structure can be manipulated JavaScript. \label{item:dataset_challenge3}
    \item Wireframes only represent structure with no content. Websites contain structure and content. Content can alter the structure i.e. two web pages may have a \lstinline{<p>} (paragraph) element but if they had different length text the size of the paragraph would be different. As such, the structure would change. \label{item:dataset_challenge4}
    \item HTML is sometimes invalid or poorly formatted, this would decrease our dataset quality. \label{item:dataset_challenge5}
    \item HTML has multiple representations of the same structure. This dilutes the quality of the dataset. \label{item:dataset_challenge6}
\end{enumerate}

These issues highlight the need for careful data processing.

\subsection{Normalisation}

\begin{figure}[H]
    \begin{subfigure}[H]{0.5\textwidth}
        \includegraphics[width=\textwidth]{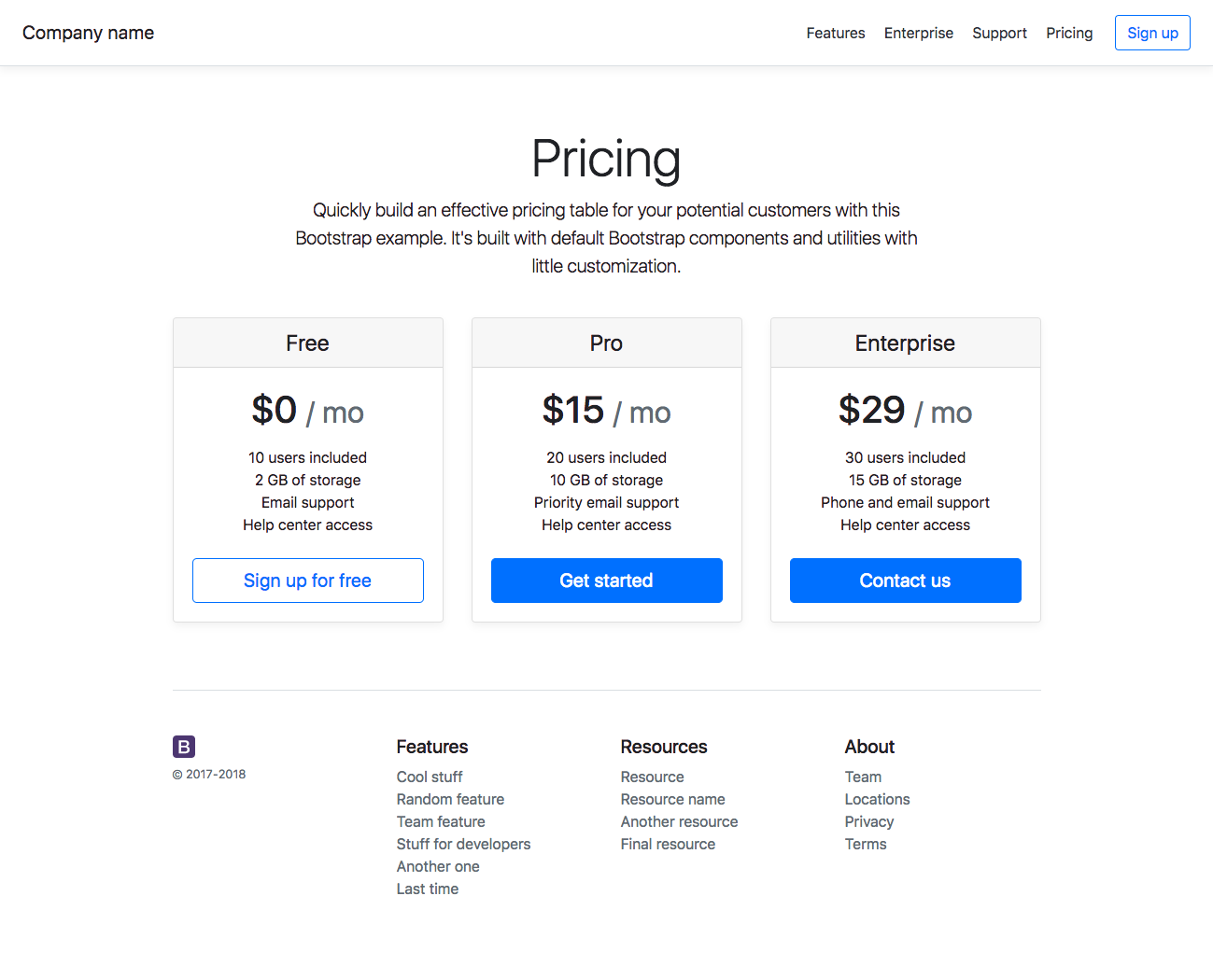}
        \caption{The original website}
    \end{subfigure}
    \begin{subfigure}[H]{0.5\textwidth}
        \includegraphics[width=\textwidth]{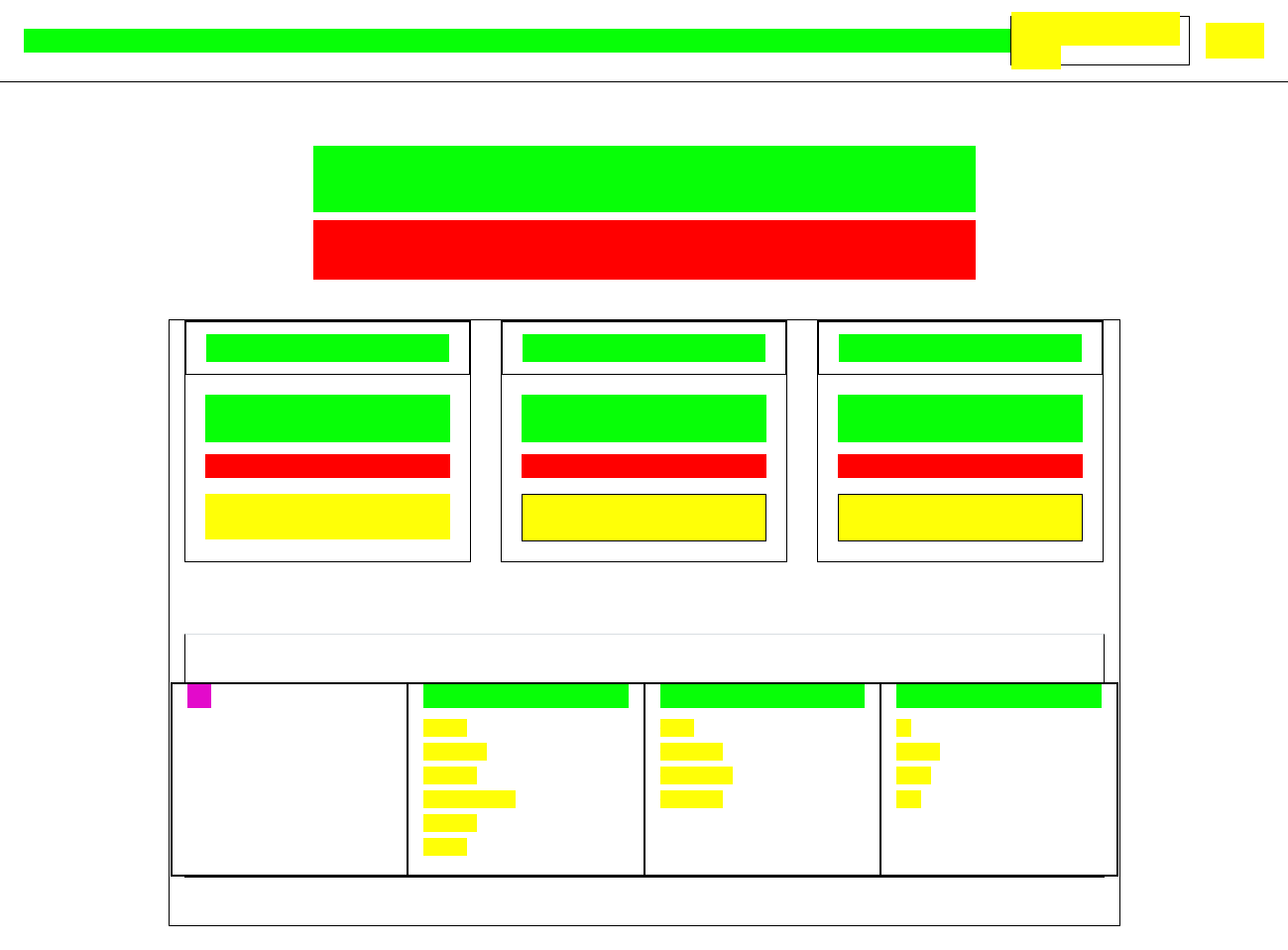}
        \caption{The normalised page}
    \end{subfigure}
    \caption{Demonstrates the original and normalised webpage. Element classes have been filled with colours; fonts, shading, and colours have been removed; the widths of elements have been maximised; JavaScript and animations have been removed. This is required to aid extraction of the website structure.}
    \label{fig:example_normalised_page}
\end{figure}

In order to mitigate the challenges stated above we developed a website normalisation process. We use PhantomJS \cite{PhantomJS}, a scriptable headless rendering engine, to apply the following transformations to normalise the webpage:

\begin{itemize}
    \item For \ref{item:dataset_challenge1} we simplified the HTML element set by merging elements into broader element classes which match wireframe elements:
        \begin{itemize}
            \item Image: \lstinline{<img>}, \lstinline{<svg>}, \lstinline{<video>}
            \item Input: text input, range inputs, text areas, sliders
            \item Button: \lstinline{<button>}, \lstinline{<a>}
            \item Paragraph: \lstinline{<p>}, \lstinline{<span>}, \lstinline{<strong>}, \lstinline{<i>}
            \item Title: \lstinline{<h1>}, \lstinline{<h2>}, \lstinline{<h3>}, \lstinline{<h4>}, \lstinline{<h5>}
        \end{itemize}
    \item To address challenge \ref{item:dataset_challenge2}, all style from elements is replaced. Shadows, borders, fonts, and colours are  removed from elements and replaced with consistent values. Elements background colours are replaced with a background color which maps to their assigned label e.g. the background colour of all image elements is replaced with the color \#FF00FF (purple). The background is replaced so as to easily identify which label an element has been assigned.
    \item \ref{item:dataset_challenge3} is addressed by disabling JavaScript and removing all animations.
    \item \ref{item:dataset_challenge4} is mitigated by setting title and paragraph elements to 100\% width and setting other elements to constant width.
\end{itemize}

\subsubsection{Structural Extraction} \label{section:dataset_tree}

Normalisation solved a number of the challenges with using real web pages. However, the challenge of poorly formatted code and multiple representations of the same structure remained an issue.

We define the structure of a webpage as the type of element, position, sizing, and hierarchy tree. We considered two approaches to extract the structure from a webpage:

\begin{itemize}
    \item Directly parsing the HTML. This approach requires many special cases as well as not solving the problem that there are multiple ways to represent the same structure. Further, correctly capturing the position of elements can sometimes be non-trivial due to CSS being able to move elements.
    \item Using computer vision to extract structure from a screenshot of the webpage. This solves the problem of multiple ways to represent structure as the layout is processed in the same way each time so two identical webpages which in HTML may have slightly different structures, would be processed identically.
\end{itemize}

We opted for the later approach. This approach involves three phases: i) screenshotting the webpage, ii) extracting elements from the screenshot, and iii) inferring structure from the extracted elements.

\begin{enumerate}[label=(\roman*)]
    \item Screenshotting the webpages. We used PhantomJS to render the webpages with AppleWebKit/538.1 website rendering engine (different engines may render pages slightly differently). Websites were normalised before screenshotting.
    \item Element extraction, from our normalisation we had already coloured specific elements in different colours and we took each screenshot and used colour filtering to isolate one type of element at a time e.g. we use a blue filter which displays only blue objects. Once each group was isolated we used contour detection to detect closed shapes, these represented our elements and we knew the types from the colours.
    \item Inferring structure. From phase (ii) we had a list of all elements with their types, positions, and sizes. However, we did not have hierarchical information. We used algorithm 1 to infer this information.
\end{enumerate}

\begin{algorithm}[H]
    \SetAlgoLined
    
    \KwData{Elements ordered by area ascending: $E={(x_0,y_0,w_0,h_0,[]),...,(x_n,y_n,w_n,h_n,[]),}$} \
    $sectioned \gets []$ \;
    \For{$element_1$ in $E$}{
        $contains \gets []$ \;
        \For{$element_2$ in $sectioned$}{
            \If{$element_1$ contains $element_2$}{
                $contains.push(element_2)$ \;
                $sectioned.remove(element_2)$ \;
            }
            $element_1.contains \gets contains$ \;
        }
        $sectioned.append \gets element_1$ \;
    }
    
    \caption{Recursive algorithm which builds tree based on bounding box hierarchy. It works by recursively grouping nested objects within one another to build a tree.}
    \label{algorithm:structural_detection}
\end{algorithm}

\subsubsection{Sketching}

\begin{figure}[H]
    \centering
    \begin{subfigure}[H]{0.3\textwidth}
       \includegraphics[width=\textwidth]{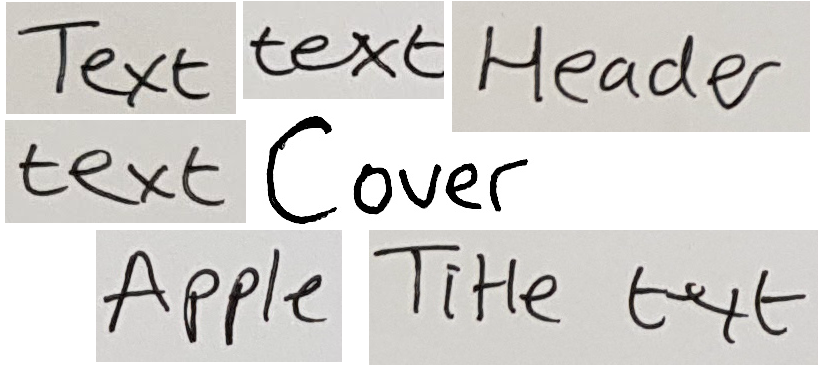}
        \caption{Title elements}
    \end{subfigure}
    \begin{subfigure}[H]{0.3\textwidth}
        \includegraphics[width=\textwidth]{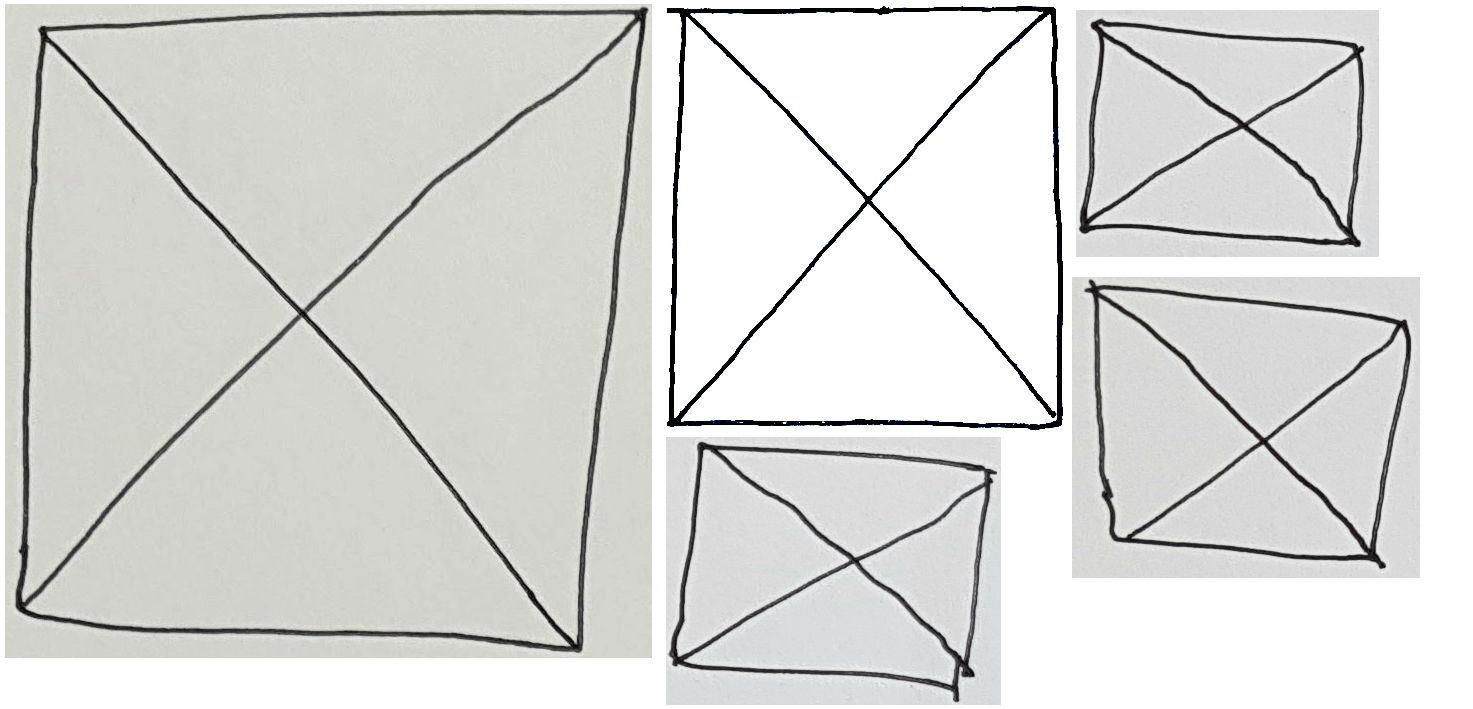}
        \caption{Image elements}
    \end{subfigure}
    \begin{subfigure}[H]{0.3\textwidth}
        \includegraphics[width=\textwidth]{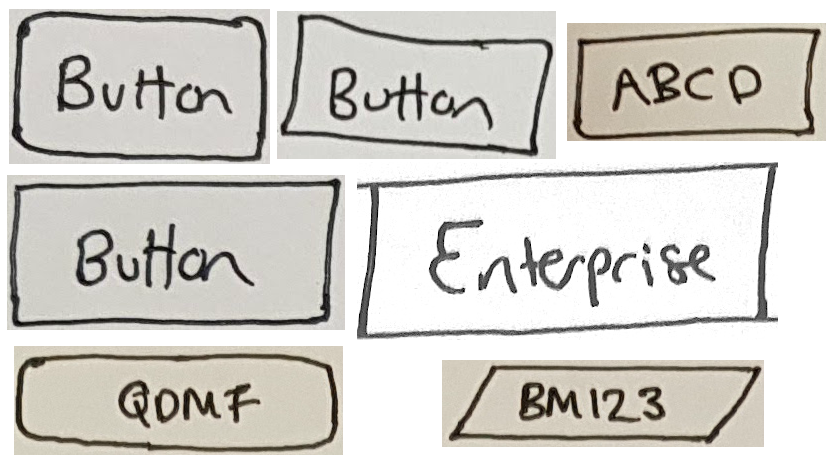}
        \caption{Button elements}
    \end{subfigure}
    \begin{subfigure}[H]{0.3\textwidth}
        \includegraphics[width=\textwidth]{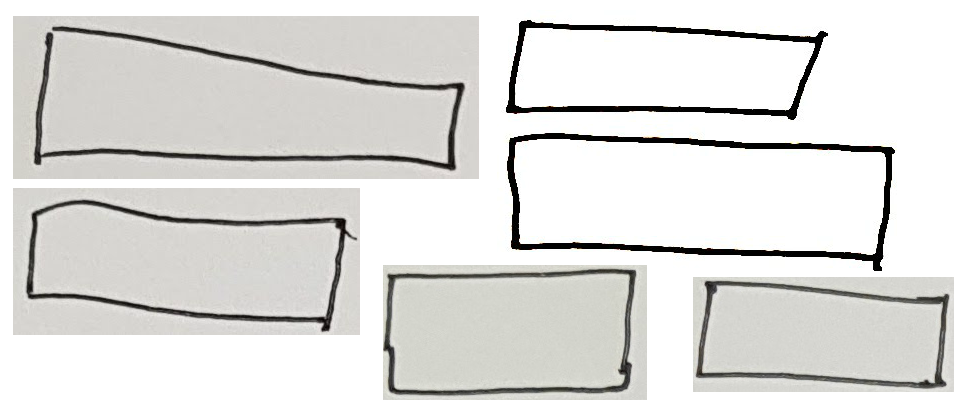}
        \caption{Input elements}
    \end{subfigure}
    \begin{subfigure}[H]{0.3\textwidth}
        \includegraphics[width=\textwidth]{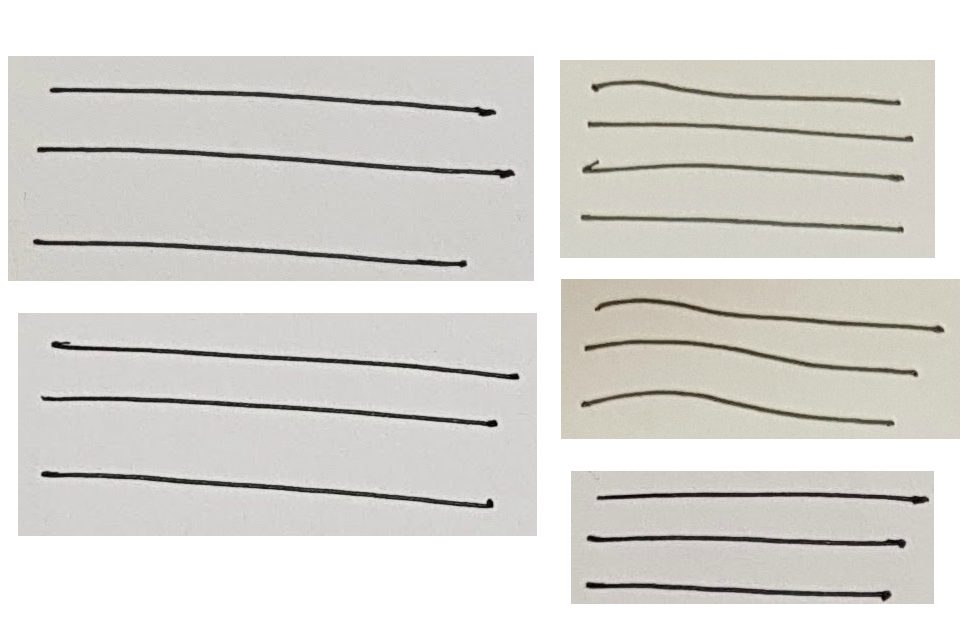}
        \caption{Paragraph elements}
    \end{subfigure}
    \caption{Samples of sketched elements from each of the five classes from our set of 90. Elements from a real website are replaced with sketched elements to create a sketched version of the webpage. }\label{fig:drawn_components}
\end{figure}

To generate a sketched version of a website we created a total of 90 individual sketches of elements, 18 for each of the five classes of elements. For each website we replaced every element with a randomly selected sketched version from the matching class. We added variation by randomly translating, scaling, and rotating elements by up to $\pm2.5\%$. We found that $\pm2.5\%$ gave similar variability to how a human would have drawn the element while not so much as to degrade the structure e.g. by overlapping elements.


When replacing elements it was important to scale the element sketches and preserve all lines. We used openCVs INTER\_AREA \cite{INTERAREA} method which resamples using pixel area relation to scale and disabled anti aliasing. Another consideration made was to not scale the widths of lines with the size of the element, as we assumed our sketches would be done in a pen/pencil with a constant thickness. To do this we applied a skeletonisation algorithm \cite{abu2013skeletonization} to all elements after they had been scaled. Skeletonisation takes a thick line and reduces it to a single pixel line. Further, text must maintain its aspect ratio in order for it to remain readable. We took this into consideration and maintained the aspect ratio of text when scaling.

\begin{figure}[H]
    \begin{subfigure}[H]{0.3\textwidth}
        \includegraphics[width=\textwidth]{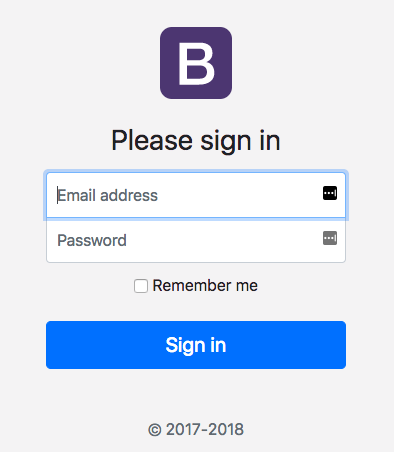}
        \caption{Original website}
        \label{fig:dataset_example_original}
    \end{subfigure}
    \begin{subfigure}[H]{0.3\textwidth}
        \includegraphics[width=\textwidth]{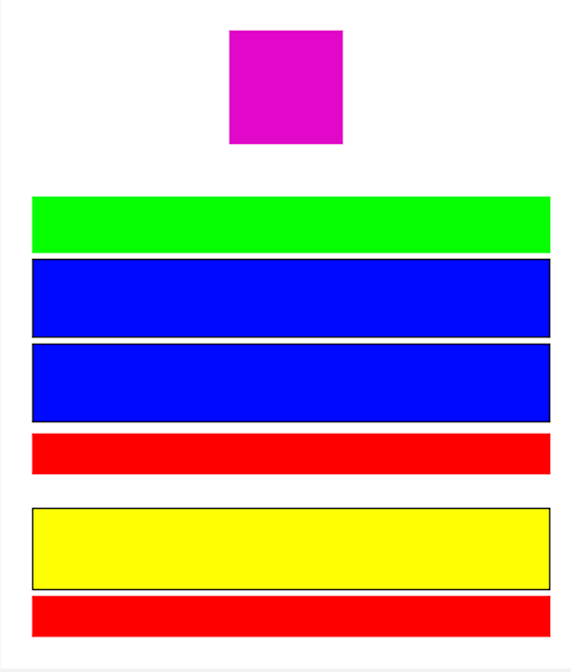}
        \caption{Normalised version}
        \label{fig:dataset_example_norm}
    \end{subfigure}
    \begin{subfigure}[H]{0.3\textwidth}
        \includegraphics[width=\textwidth]{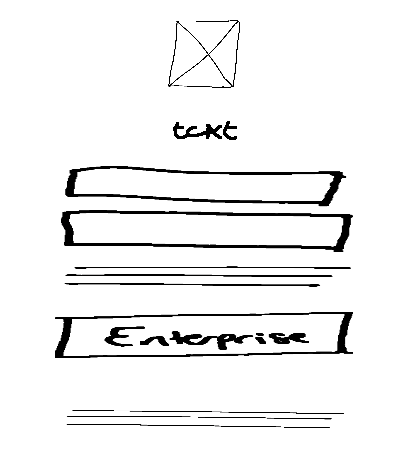}
        \caption{Sketched version}
        \label{fig:dataset_example_drawn}
    \end{subfigure}
    \caption{The dataset processing steps. The original website (a) is normalised (b) which removes style and maps HTML elements into our wireframe element classes. The structure is then extracted by detecting the colours and using algorithm 1. From the structure a sketched version of the website is created by replacing elements with a random sketched version. Resulting in (c).}
    \label{fig:drawn_website}
\end{figure}

\subsection{Dataset}

\begin{figure}[H]
    \centering
    \begin{subfigure}[H]{0.45\textwidth}
        \includegraphics[width=\textwidth]{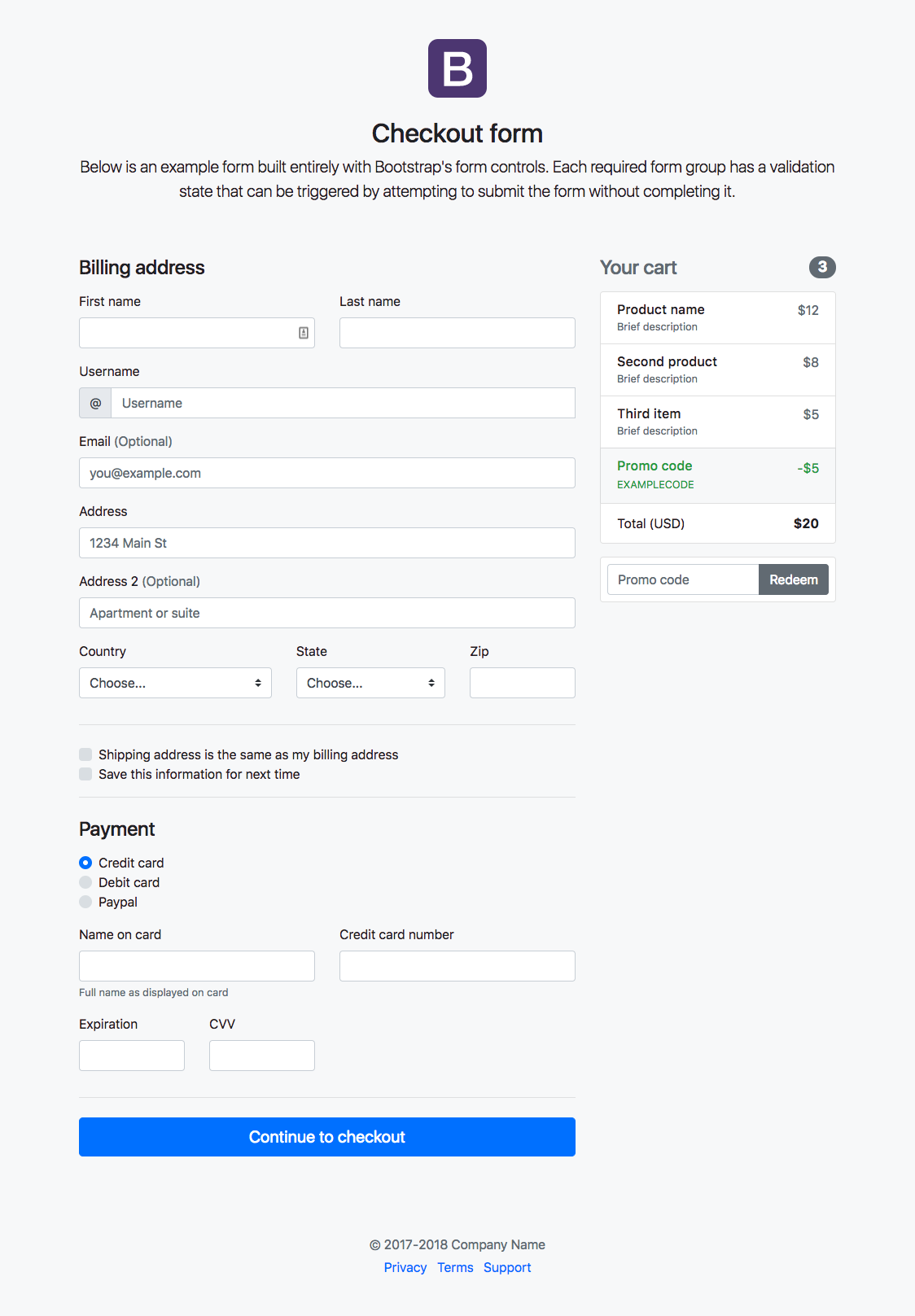}
        \caption{Screenshot of http://getbootstrap.com/docs/4.0/examples/checkout/}
    \end{subfigure}
    \begin{subfigure}[H]{0.45\textwidth}
        \includegraphics[width=\textwidth]{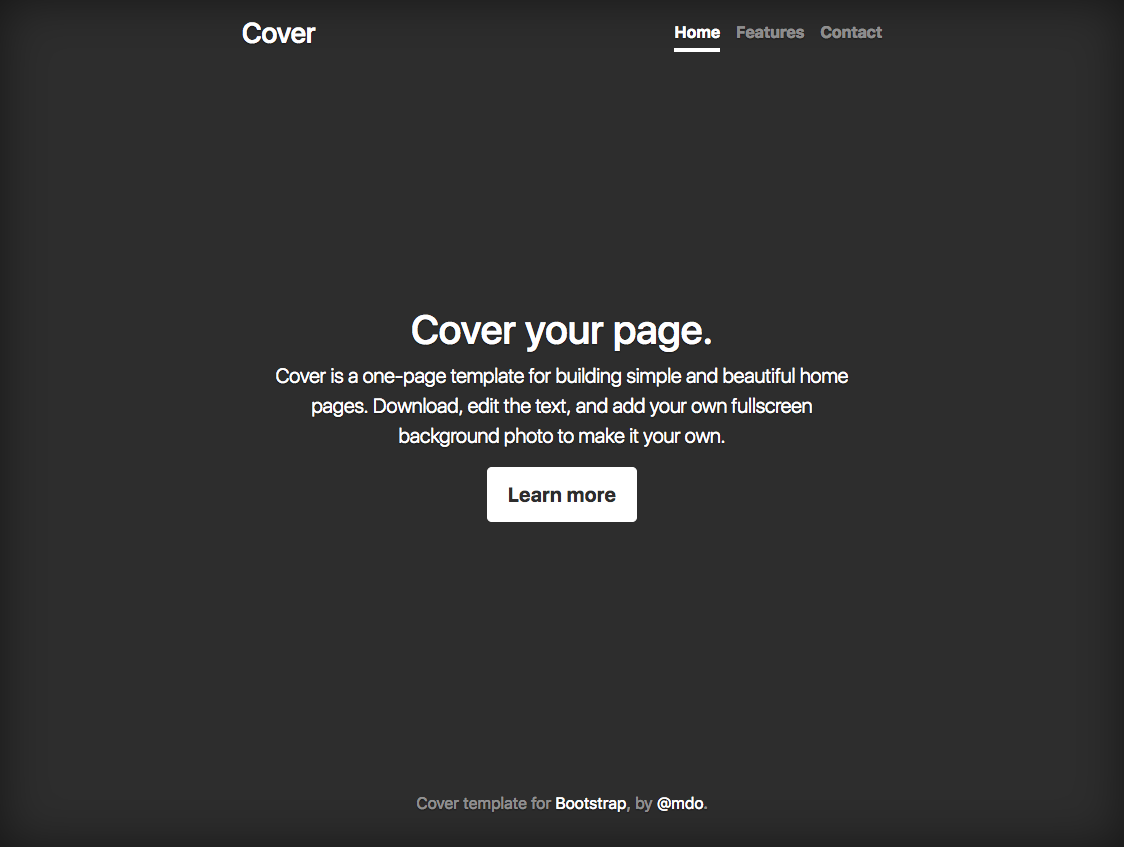}
        \caption{Screenshot of http://getbootstrap.com/docs/4.0/examples/cover/}
    \end{subfigure}
    \caption{Sample of two webpages collected.}
    \label{fig:dataset_example_webpages}
\end{figure}

Challenge \ref{item:dataset_challenge5} - HTML is sometimes invalid or poorly formatted - was partially solved by structural extracting using computer vision. However, in order to maximise the quality of the website code we collect Bootstrap website templates. Bootstrap \cite{Bootstrap} is a popular website style and structural framework. We targeted Bootstrap templates as it provided a more consistent framework which would limit the number of special cases which our normalisation process would have to deal with. Further, Bootstrap enforces best practices such as using the semantic \lstinline{<header>} and \lstinline{<footer>} tags as well as ``row" and ``container" classes which allows us to easily categorise the structure of container elements. Bootstrap does not limit the structure of a webpage and therefore a dataset of Bootstrap templates should still remain general.

\begin{table}[H]
\begin{center}
\begin{tabular}{ |c| } 
\hline
URL \\
\hline
https://blackrockdigital.github.io/startbootstrap-landing-page/ \\
http://getbootstrap.com/docs/4.0/examples/album/ \\
http://getbootstrap.com/docs/4.0/examples/pricing/ \\
http://getbootstrap.com/docs/4.0/examples/checkout/ \\
http://getbootstrap.com/docs/4.0/examples/cover/ \\
http://getbootstrap.com/docs/4.0/examples/carousel/ \\
http://getbootstrap.com/docs/4.0/examples/blog/ \\
http://getbootstrap.com/docs/4.0/examples/sign-in/ \\
\hline
\end{tabular}
\end{center}
\caption{A sample of URLs from the 1750 websites used to create our dataset.}
\label{table:dataset_urls}
\end{table}

Generally, more data leads to higher performance in machine learning models \cite{halevy2009unreasonable}. We curated a list of 1750 URLs to collect. 1750 was chosen as it was the same as pix2code's dataset which was the closest research to our own.

We downloaded and inlined (to remove external file requirements by placing everything in the HTML) the URLs. This was done to reduce traffic to the site by making a local copy to operate on. Each website was then normalised, the structure extracted, and then sketched.

\section{Framework} \label{section:framework}
\begin{figure}[H]
    \centering
    \includegraphics[width=0.8\textwidth]{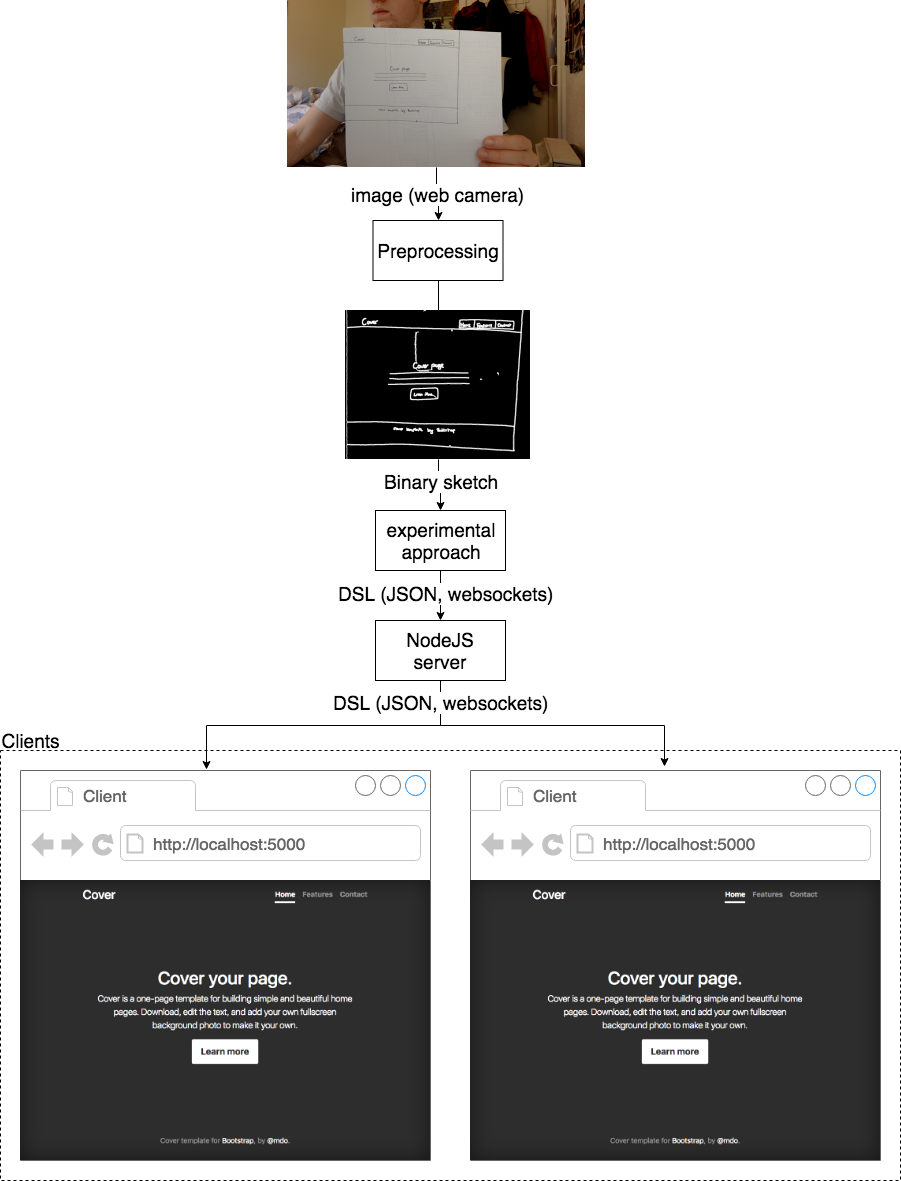}
    \caption{The framework performs preprocessing turning an image from a camera into a binary sketch. The sketch is passed through either experimental approach and the resulting generated code is passed to a NodeJS web server which then broadcasts the code to all clients. Clients translate the code into HTML client side.}
    \label{fig:server}
\end{figure}

To make our methods easier to demonstrate, and to allow the experimental approaches to solely focus on the challenge of translating an image into code, we have designed a framework to perform the pre and post processing required to translate an image taken from a camera through to a live updating website which renders the generated code. The framework is intended to be general and allow future experimental approaches to be incorporated.

We expect a sketch to be drawn on a white medium in a dark marker, for example a pen on paper or a marker on a whiteboard.

Note that this is required to make our program operational. However, we focus our evaluation purely on the experimental methods.

\subsection{Preprocessing}
\begin{figure}[H]
    \centering
    \includegraphics[width=\textwidth]{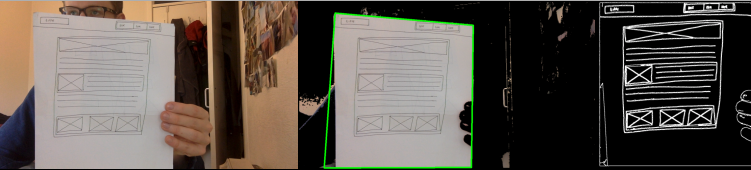}
    \caption{Example of crop to sketch method. This process takes a raw image from a camera and returns a black and white image of just the sketch. From left to right: the original input image, the image with all colours except white filtered and candidate contours detected, and the deskewed edge map of the selected region.}
    \label{fig:preprocessing}
\end{figure}

Preprocessing is required to translate an image from a camera into an image which can be fed into the experimental method. Due to the positioning of the camera or lighting conditions the raw image must be cleaned up before it can be processed.

The main challenges are:

\begin{enumerate}[label=(\roman*)]
    \item The image may not fill the entire frame, as such the background must be removed. \label{item:framework_challenge_1}
    \item The paper may be skewed or rotated. \label{item:framework_challenge_2}
    \item The image may contain noise or alterations due to lighting. \label{item:framework_challenge_3}
\end{enumerate}

We address each of these challenges in turn:

\begin{itemize}
    \item To address \ref{item:framework_challenge_1} we detect the paper in the image and crop to it. As our requirements state that the medium must be white, to detect the paper we converted the image to HSV and used threshold filtering to remove all colours except white. This process reduces the background noise considerably. As the paper contains the sketch in a dark marker, these pixels were filtered by the threshold filtering. As such, to fill the gaps left by the sketch we applied a large median blur. We then applied Canny edge detection and dilate the edge map to close small gaps between the edges. Finally, we applied contour detection and found the largest contour with approximately four sides (as we assume the medium is four sided e.g. paper or a white board).
    \item To address \ref{item:framework_challenge_2} we used perspective warping \cite{wolberg1994digital} to unwrap the four corners of the contour found and map them to an equivalent rectangle.
    \item To address \ref{item:framework_challenge_3} we applied a medium blur to the unwrapped cropped image to reduce noise. We then ran Canny edge detection and dilated the edge map to close small gaps between lines. 
\end{itemize}

The result of the post processing is an unskewed binary edge map of the sketch. This is fed into the experimental method for processing.

\subsection{Post-processing}

The output of the experimental method is code which represents the structure of the wireframe. There are three post-processing steps: 1) distribute the generated code from the experimental approaches to clients, 2) translate the generated code into HTML, and 3) live update the website.

\begin{figure}[H]
    \centering
    \begin{lstlisting}
    {
        'type': <the element type>,
        'x': <relative x position>,
        'y': <relative y position>,
        'width': <the element width relative to the page width>,
        'height': <the element height relative to the page height>,
        'left_padding': <relative left padding>
        'top_padding': <relative top padding>
        'contains': [<elements>],
    }
    \end{lstlisting}
    \caption{JSON tree-like structure used to represent structure of a wireframe. This can be directly translated into HTML.}
    \label{fig:DSL}
\end{figure}

Note that the approaches do not generate HTML directly but instead we use an intermediary \textit{domain specific language} (DSL) which represents the tree like layout representing the structure of the wireframe. A DSL is used beacuse the representation of the structure of the wireframe does not directly match to HTML as we use only five element classes. As such, a translation step is required to convert our DSL into HTML, we encapsulate this process in our framework so that our experimental methods can focus purely on the task of translating an image to code. We use \textit{JavaScript object notation} (JSON) as a carrier syntax for our DSL as seen in figure \ref{fig:DSL}.

Step 1) is achieved by the generated DSL from the experimental approach being sent over WebSockets \cite{WebSockets} to a NodeJS \cite{Nodejs} web server. WebSockets are a protocol providing full-duplex communication channels over a single TCP connection. The web server then broadcasts the DSL over WebScokets to connected clients. Clients connect via visiting the servers address in a web browser. By using a web server to broadcast the DSL we allow multiple clients to connect and view the generated webpage. This is a usability feature to aid collaboration. By using WebSockets we can send updates without the client refreshing their browser this achieves step 3).

Step 2) is achieved by each client using JavaScript to translate the DSL into HTML and update the DOM.

Further usability features have been added: drag and drop image replacement, text replacement, and click to change colour. These allows a page to be quickly styled.

\section{Approach 1: Classical Computer Vision} \label{section:approach1}
In this section we detail our baseline method which primarily uses computer vision to solve the task of converting an image of a sketch into code. This approach involves four key phases:

\begin{itemize}
    \item Element detection - use computer vision to detect and classify the position, sizes and types of all elements from the sketch. Required for reproducing equivalent elements in HTML.
    \item Structural detection - produce the hierarchical tree from the list of all elements. Required for correctly reproducing the element tree in HTML.
    \item Container classification - classify the types of container structures e.g. headers and footers. Required as aids producing the correct structure as well as creating semantically correct and human readable code.
    \item Layout normalisation - correct for human errors in the sketching process such as misaligned elements. Required to produce correct element tree in HTML.
\end{itemize}

\subsection{Element detection} \label{section:element_detection}

\begin{figure}[H]
    \centering
    \includegraphics[width=0.5\textwidth]{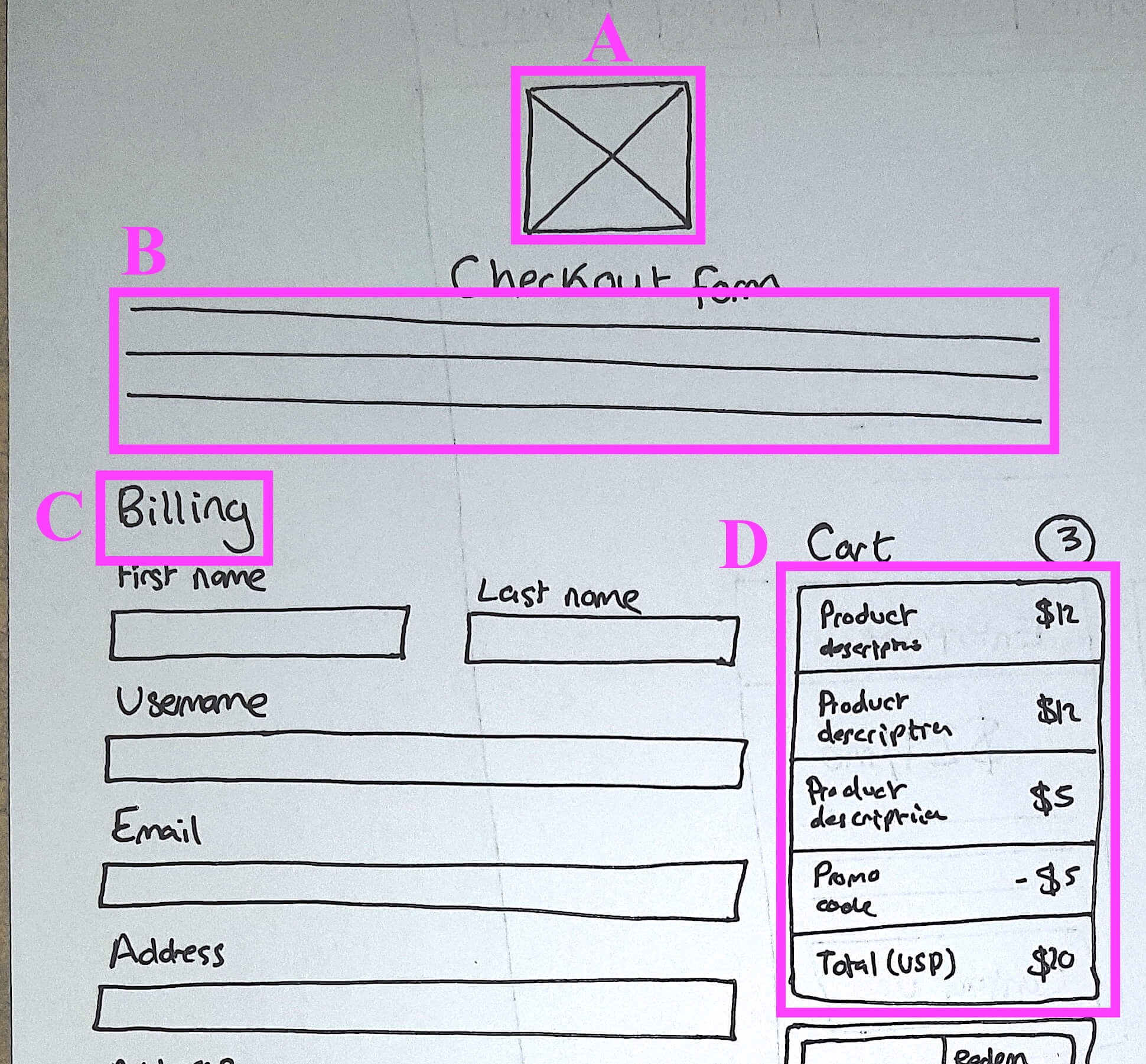}
    \caption{Partially labeled elements on wireframe: (a) an image, (b) a paragraph, (c) a title, and (d) a stack container. Element detection aims to classify and detect the bounding box of all elements in an image.}
    \label{fig:atomic_components_labeled}
\end{figure}

The aim of this phase is to take an image and produce a list of elements with their bounding box (position and size), and element type. As explained in section \ref{section:background} we will base classification on the five element classes: images, paragraphs, titles, inputs, and buttons. From the framework described in section \ref{section:framework} we are fed a preprocessed binary edge map as our input we assume this is the input in all of the methods described below.

For approach 1 we restrict our focus to using computer vision techniques and look at deep learning techniques in section \ref{section:approach2}. We considered two approaches to classify elements:

\begin{itemize}
    \item Extract features and use machine learning. Apply Harris corner detection \cite{Harris} or key point detectors such as SIFT \cite{SIFT} and train a machine learning model to classify elements based on common features.
    \item Use contour detection to manipulating and filter the image. Use contour detection to detect shapes in the sketch. Devise a filtering process to repeatedly filter contours based on properties such as number of sides, extent, solidity, and aspect ratio. In order to classify elements.
\end{itemize}

We opted for the latter approach as we believed that the former approach would require a large dataset of sketched elements in order to generalise well. From our dataset we only had 90 elements, and it would be very time consuming to make enough for a large dataset. Note that we use this approach for paragraphs, inputs, containers, and buttons. For text we use a text detection method.

\subsubsection{Images}

\begin{figure}[H]
    \centering
    \includegraphics[width=0.5\textwidth]{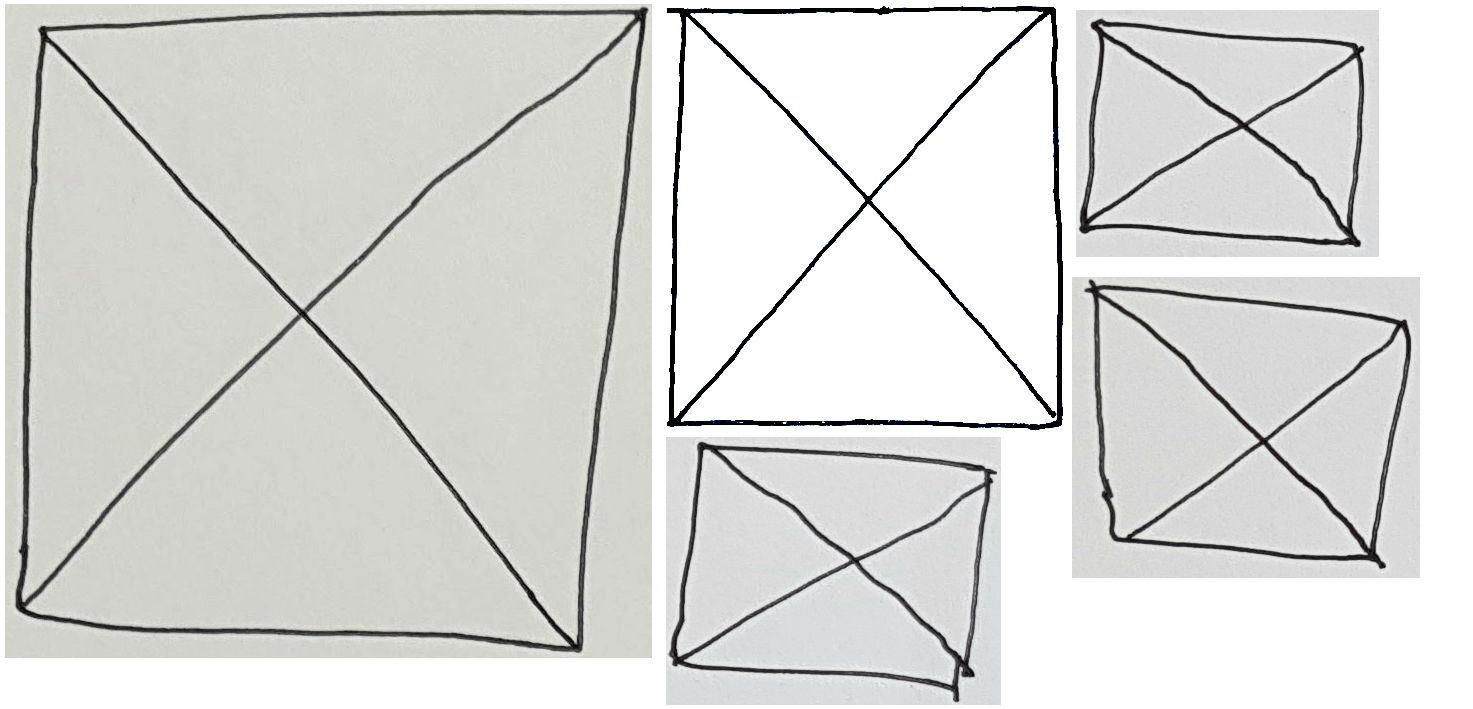}
    \caption{Sample of wireframe image elements. Elements are roughly rectangular with a cross through the middle from each corner. We used the fact that the box with a cross creates four triangles in order to classify this element.}
    \label{fig:approach1_wireframe_images}
\end{figure}

\begin{figure}[H]
    \begin{subfigure}[H]{0.3\textwidth}
        \includegraphics[width=\textwidth]{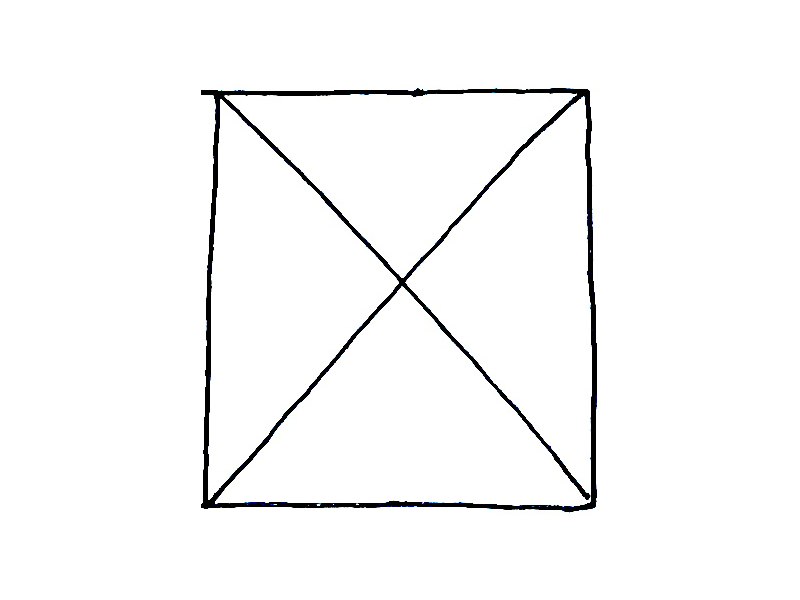}
        \caption{The image element we want to detect}
    \end{subfigure}
    \begin{subfigure}[H]{0.3\textwidth}
        \includegraphics[width=\textwidth]{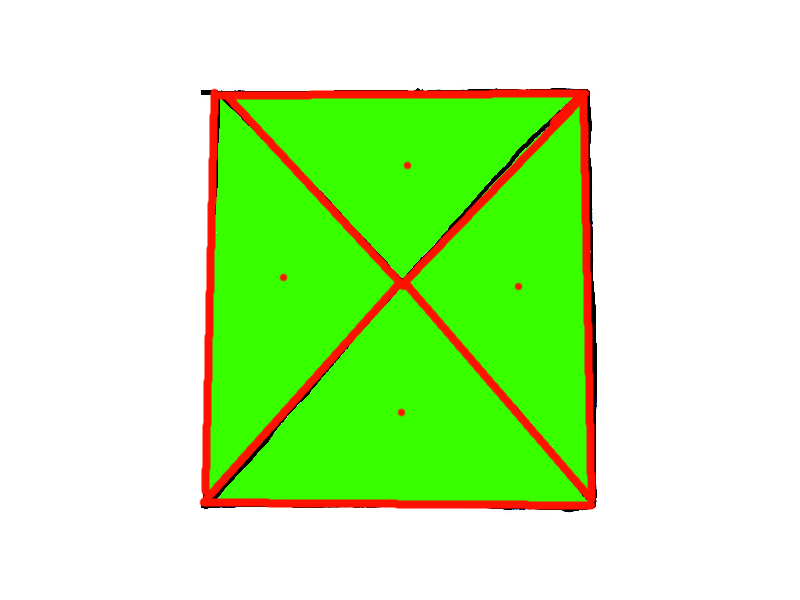}
        \caption{The 4 triangles detected}
    \end{subfigure}
    \begin{subfigure}[H]{0.3\textwidth}
        \includegraphics[width=\textwidth]{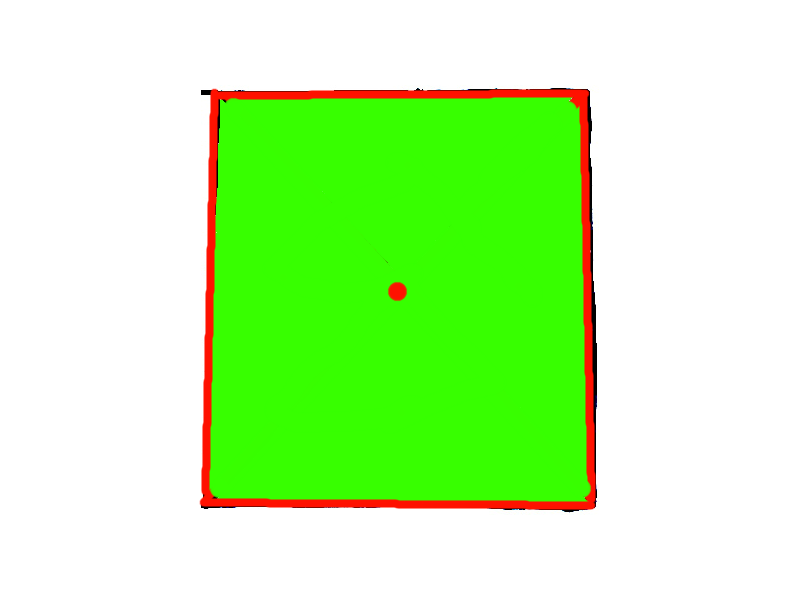}
        \caption{The rectangular contour contains four triangles and so it is classified as an image}
    \end{subfigure}
    \label{fig:approach1_image_stages}
    \caption{To detect and classify an image element we initially detect rectangular contours, we then detect triangular contours, if a rectangular contour contains three or four triangles we classify it as an image.}
\end{figure}

We identified that the symbol is made up of a box with four triangles inside and used these features to build a classifier. Firstly, we used contour detection to find all contours with approximately four sides and removed all other shapes. We then ran another round of contour detection and collected all triangular contours. If a rectangle had between three and four triangles, and those triangles occupied the majority of the shape we consider it to be an image symbol. 

\subsubsection{Paragraphs}

\begin{figure}[H]
    \centering
    \includegraphics[width=0.5\textwidth]{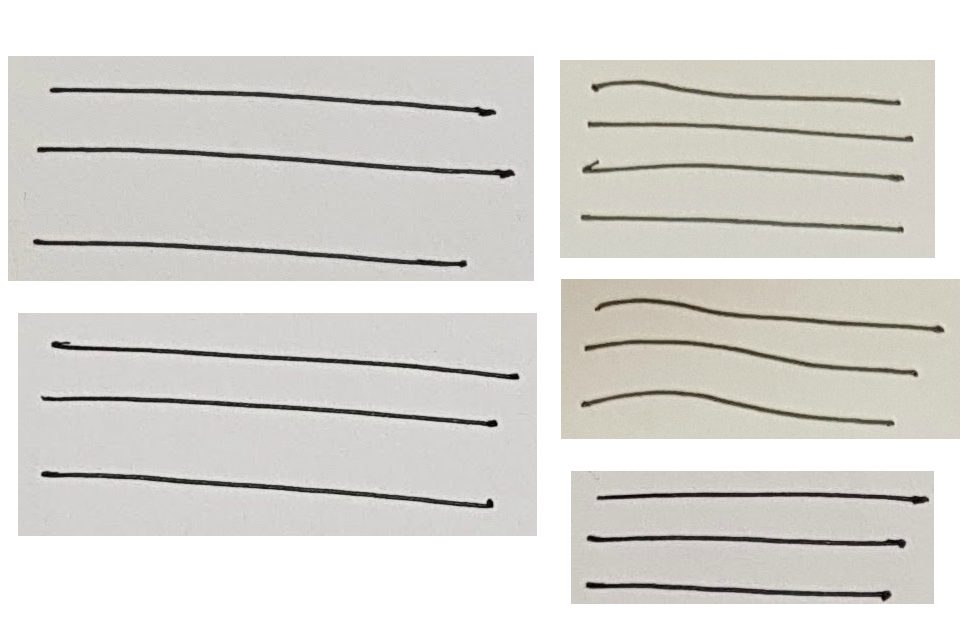}
    \caption{A sample of wireframe paragraph elements. Elements consist of three or more mostly straight horizontal lines of roughly equal size. These are key characteristics we use to detect and classify these shapes.}
    \label{fig:approach1_wireframe_paragraphs}
\end{figure}

\begin{figure}[H]
    \centering
    \includegraphics[width=\textwidth]{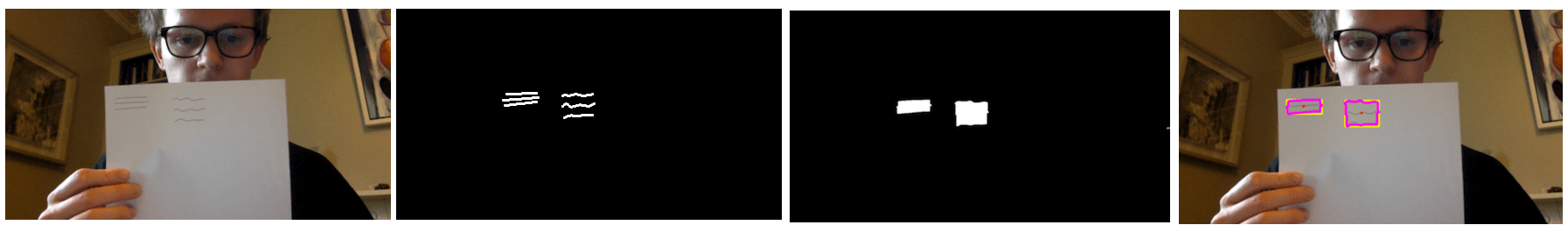}
    \caption{Paragraph detection method. From left to right: original image with two paragraphs, lines detected, lines with large close operation, contour detection filtering non-rectangles and removes if $< 3$ lines.}
    \label{fig:approach1_wireframe_paragraphs}
\end{figure}

The paragraph symbol is made up of three or more lines (straight or wavy) of approximately equal length. We applied a similar method as with images to classify these shapes. We apply contour detection and filter out all closed shapes - a closed shape is a shape which encloses an area. This will leave only marks such as lines which do not enclose an area. From these we filter out all shapes which have an aspect ratio $> 0.15$ as we are looking for long lines, we found that $0.15$ produced the fewest false positives by experimentation on our 18 paragraph examples (see section \ref{section:background}). Our image now only contains the lines which passed this filtering and we apply a dilation operation to fill the gaps between close lines. We then apply another contour detection and filter out non-rectangles, for each of these we count the number of lines we detected from the previous pass, if this is $\geq 3$ the detected shape is a paragraph.

\subsubsection{Inputs}

\begin{figure}[H]
    \centering
    \includegraphics[width=0.5\textwidth]{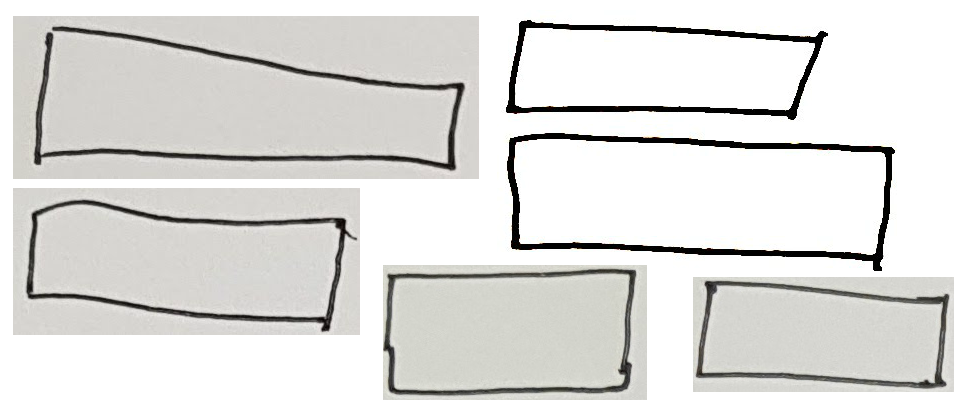}
    \caption{Sample of wireframe input elements. They consist of a closed rectangle with an aspect ratio $<0.3$. We use these features to detect and classify.}
    \label{fig:approach1_wireframe_input}
\end{figure}

An input is simply a closed rectangle containing nothing else - they are longer than they are tall. We apply contour detection and classify all rectangular contours with an aspect ratio $<0.3$ and which contain no other shapes to be input elements.

\subsubsection{Containers}

In this phase we only detect containers, we classify containers in section \ref{section:container_classification}. Containers (see figure \ref{fig:atomic_components_labeled}) are rectangles which contain other shapes and are used to imply structures such as rows. We apply contour detection filtering only on rectangular shapes, we consider a contour a container if it contains elements which we have already classified using all other methods. 

\subsubsection{Titles}

\begin{figure}[H]
    \centering
    \includegraphics[width=0.5\textwidth]{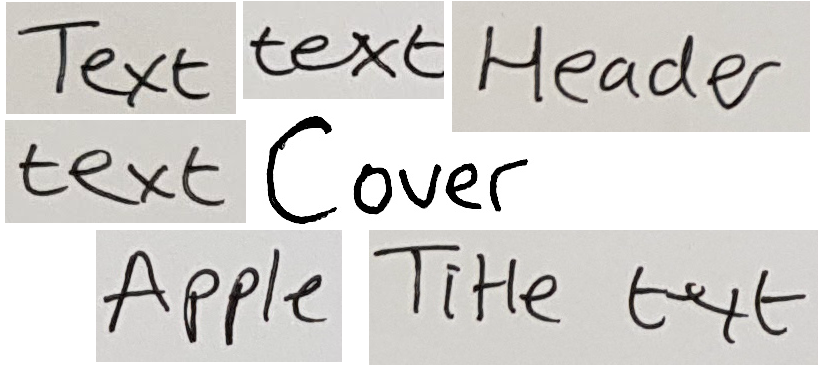}
    \caption{Sample of wireframe title elements. Title elements consist of text with no boxes around it.}
    \label{fig:approach1_wireframe_title}
\end{figure}

Note that text written on paper is separate to paragraphs - in wireframes paragraphs show large blocks of text while actual text is used to represent titles.

All of the other methods primarily used contour detection to detect and classify elements but with text this becomes more difficult. Note that we are only concerned with text detection and not recognition i.e. the position and size of the text and not its content.

We used stroke width transform (see section \ref{section:background}) an off the shelf algorithm designed to work in real world scenes. As we had a consistent context we fine tuned SWT's parameters to improve accuracy and speed for our specific context of dark text on a 2D surface.

Major changes include:
\begin{itemize}
    \item An initial pass using contour detection removing non leaf contours. This resulted in a significantly smaller number of boxes to run SWT over which improved its speed.
    \item SWT searches for both dark text on lighter background and light text on darker background. As our images consisted of dark text on a lighter background we only performed a single pass thus reducing the running time by half.
    \item We found SWT detecting false positives by detecting the cross in wireframe image elements as "x"s. We added an additional post-processing step to remove detections with an aspect ratio $>0.8$. We found the $0.8$ threshold produced the highest detection accuracy on our 18 wireframe title samples.
\end{itemize}

\subsubsection{Buttons}

\begin{figure}[H]
    \centering
    \includegraphics[width=0.5\textwidth]{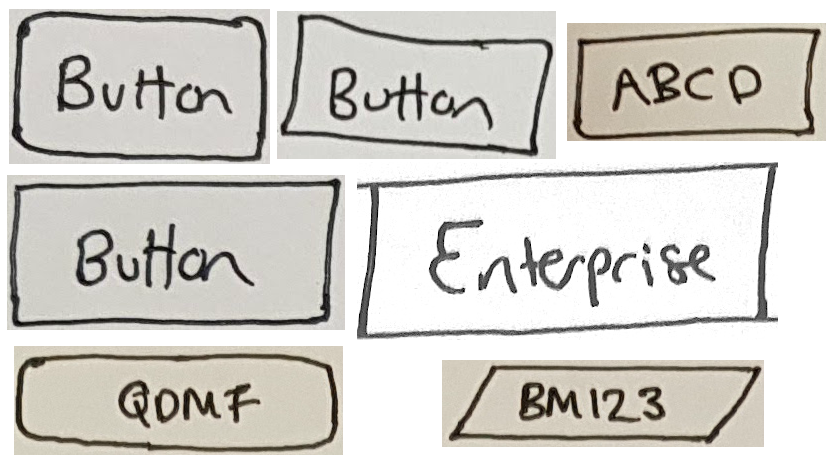}
    \caption{A sample of wireframe button elements. A button element consists of text closely surrounded by a a rectangular box with a similar aspect ratio. We use these features to detect and classify buttons.}
    \label{fig:approach1_wireframe_button}
\end{figure}

Buttons are represented in wireframes by text enclosed by a rectangular container. We took each title element which had already been detected and reclassified it as a button if it was enclosed by a rectangle which was up to 50\% larger than the text and if the container contained nothing else. We found that 50\% was the threshold which produced the highest accuracy through experimentation on our 18 button sketches.

\subsection{Structural detection} 

The previous phase parsed the input image and returned a list of all elements with their labels, positions, and sizes. In this phase we take this list and infer the hierarchical tree structure. We used the recursive algorithm 1 to perform this process.

\subsection{Container classification} \label{section:container_classification}

In this phase we take the hierarchical tree from the previous phase and classify container elements. While HTML uses a number of different tags to represent containers, such as \lstinline{<div>} or \lstinline{<span>}, we have taken inspiration from Bootstrap \cite{Bootstrap} and use types which have more semantic meaning. We classify five classes of elements:

\begin{itemize}
    \item Rows - horizontally stacked elements.
    \item Stacks - vertically stacked elements.
    \item Forms - a container containing associated inputs e.g. a contact form or search bar.
    \item Headers - a common container often at the top of a webpage containing the page title and navigation elements.
    \item Footers - a common container often at the bottom of a webpage containing contact information and other web navigation links.
\end{itemize}

We identified that a container can be classified based on structural features. Structural features include the containers bounding box (size and position) relative to the page, and the element types of sub elements. It was important to use relative positions and sizes so that they can be compared between wireframes with different dimensions. For example, a header container can be classified by the fact that it is usually near the top, spanning the entire width of the wireframe, and mostly contains buttons (links).

We considered two methods to classify based on these features:

\begin{itemize}
    \item Manually building a model. This would involve studying each container and understanding which specific thresholds of features correspond to that container type. This process is time consuming as it involves a lot of trial and error. 
    \item Using machine learning to build a model. This would involve training a machine learning model to learn which features correspond to which container type based on training data. This process requires a dataset of containers and its performance is based on the quality of the dataset.
\end{itemize}

We opted for the second method, using machine learning.

Rather than engineering features and adding human judgment, we use the raw features gathered from the previous rounds of processing i.e. the x, y, width, and height of the element as well as a flattened list of the element types of all sub elements contained. The sub elements are ordered consistently, left to right, top to bottom, smallest area to largest. Further, we cap this list at the first 50 elements or pad with null values to bring the length up to 50. As our input images may be variably sized we converted the raw pixel position and size values into values relative to the size of the page, e.g. rather than the width being 200 pixels, we say it is 80\% of total page width.

\begin{figure}[H]
    \begin{subfigure}[H]{0.3\textwidth}
        \includegraphics[width=\textwidth]{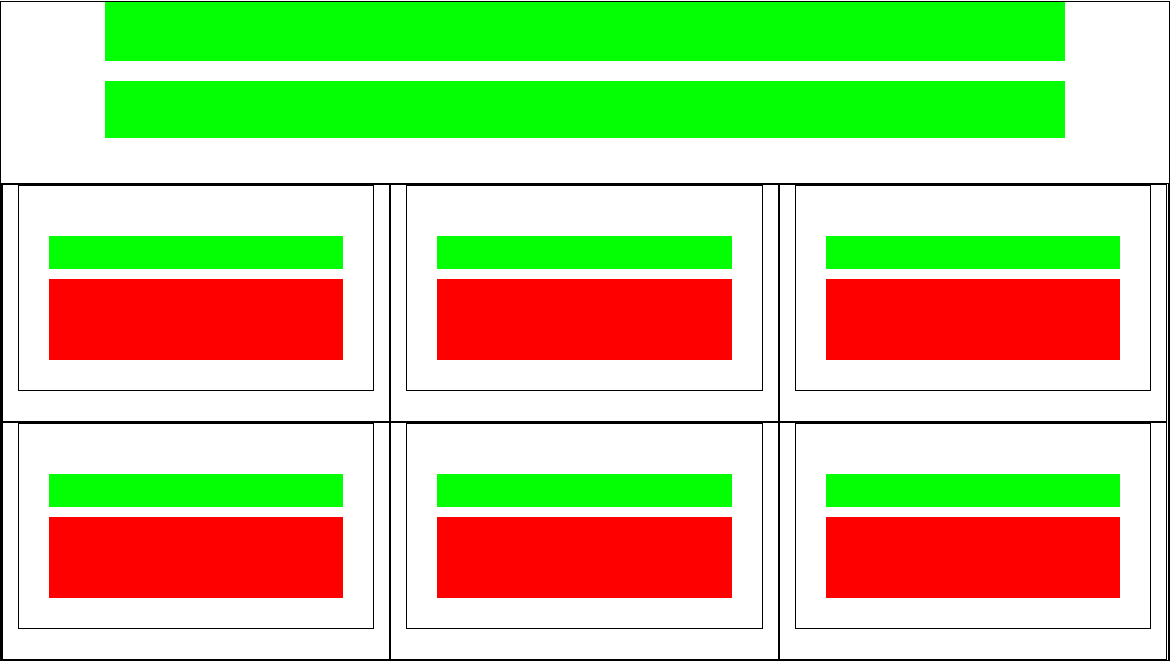}
        \caption{A stack}
    \end{subfigure}
    \begin{subfigure}[H]{0.3\textwidth}
        \includegraphics[width=\textwidth]{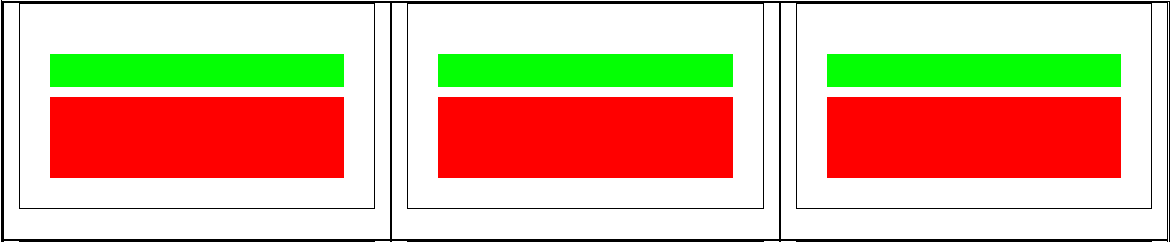}
        \caption{A row}
    \end{subfigure}
    \begin{subfigure}[H]{0.3\textwidth}
        \includegraphics[width=\textwidth]{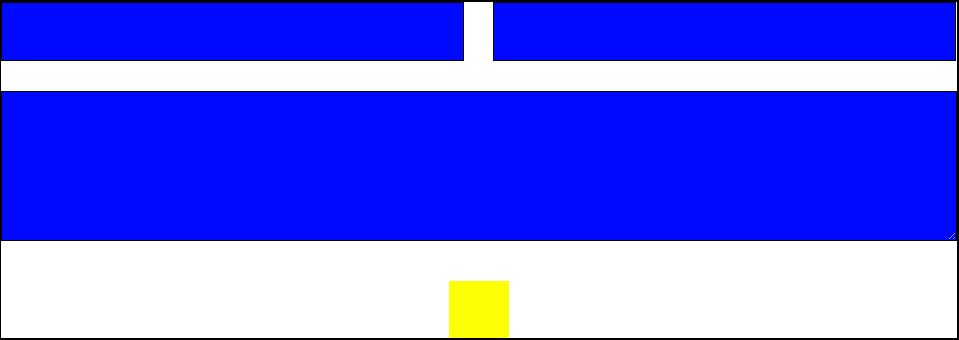}
        \caption{A form}
    \end{subfigure}
    \begin{subfigure}[H]{0.3\textwidth}
        \includegraphics[width=\textwidth]{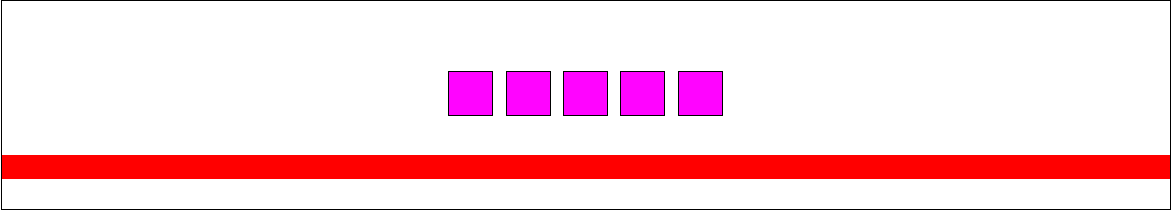}
        \caption{A footer}
    \end{subfigure}
    \begin{subfigure}[H]{0.3\textwidth}
        \includegraphics[width=\textwidth]{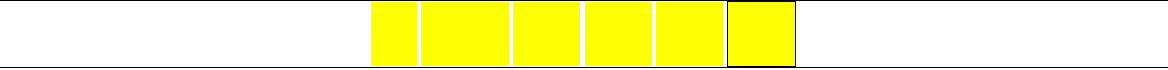}
        \caption{A header}
    \end{subfigure}
    \label{fig:example_dataset_selectors}
    \caption{A sample of normalised containers extracted from our dataset and used to train our container classifier.}
\end{figure}

\begin{table}[h!]
    \centering
    \begin{tabular}{ |c|c| } 
     \hline
     \textbf{Container class} & \textbf{Number of containers} \\
     \hline
     Stack & 4192 \\ 
     \hline
     Row & 4830 \\ 
     \hline
     Form & 562\\ 
     \hline
     Footer & 1459 \\ 
     \hline
     Header & 1388 \\ 
     \hline
    \end{tabular}
    \caption{Number of containers extrated from our dataset per container class.}
\label{table:dataset_containers}
\end{table}

We modified our existing dataset (section \ref{section:dataset}) and extracted containers from each webpage using PhantomJS. Table \ref{table:dataset_containers} shows the number of element classes, figure \ref{fig:example_dataset_selectors} shows example normalised extracted containers.
Our dataset 

\begin{figure}[H]
    \centering
    \includegraphics[width=0.75\textwidth]{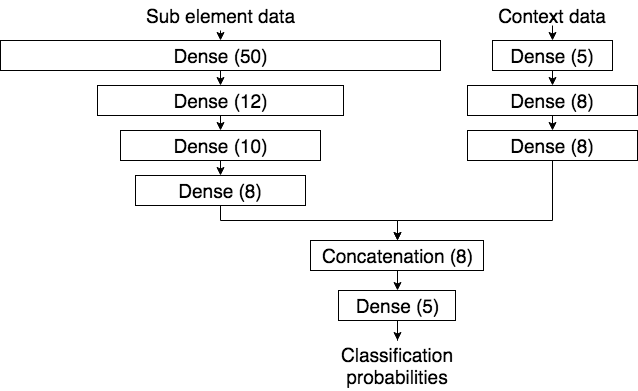}
    \caption{MLP model architecture}
    \label{fig:approah1_MLP_model}
\end{figure}

We used the common techniques \cite{lecun1998efficient} and trial and error to tune our hyperparameters to our dataset based on maximising performance on our test set.

Our model consists of two combined MLPs. One MLP is trained and learns to classify from the x, y, width, and height of the container. The other is trained and learns to classify from the element types of sub elements. We use one hot encoding \cite{nasrabadi2007pattern} to perform binarization of the categorical element types. A final MLP takes the concatenated result of both of these and produces the final classification. This model was designed to allow the network to take full advantage of both inputs. Figure \ref{fig:approah1_MLP_model} shows the model.

For hidden layers we considered both the Tanh \cite{nasrabadi2007pattern} and ReLU \cite{nair2010rectified} activation functions. We found that ReLU gave better performance. For our output layer we used softmax \cite{nasrabadi2007pattern} to give element class probabilities. We trained with a binary cross entropy loss function using the Adam optimiser \cite{DBLP:journals/corr/KingmaB14}, optimising for accuracy. We chose adam as it works well in practice and outperforms other Adaptive techniques \cite{DBLP:journals/corr/Ruder16}.

\begin{figure}[H]
    \centering
    \includegraphics[width=0.75\textwidth]{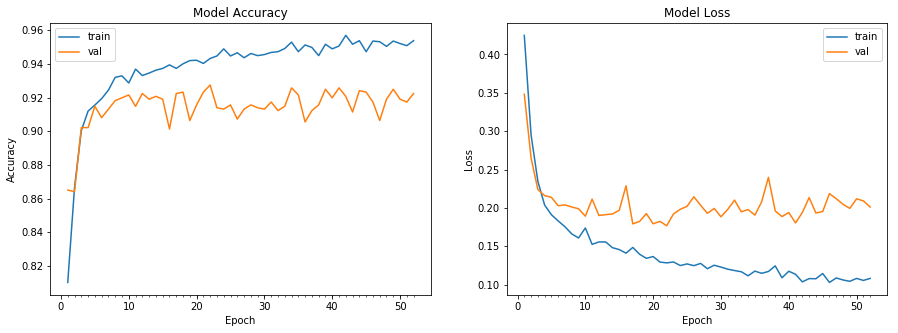}
    \caption{We used early stopping to stop training when our validation accuracy decreases and begins to drop.}
    \label{fig:approah1_MLP_model}
\end{figure}

We use early stopping to prevent overfitting and stop when our validation accuracy converges and begins to drop \cite{PIOTROWSKI201397}. We used containers from 1250 samples from our dataset with 250 samples reserved for testing and 250 for validation. After 10 epochs we achieved an accuracy score of 0.920 on the test set.

\subsection{Layout normalisation}

From the previous phases we have a tree hierarchy of elements. All leaf nodes have element types and all branch nodes have container types. However, sketched wireframes often contain human errors, types of errors are:

\begin{itemize}
    \item Rotation - a shape is drawn slightly rotated, when in fact it was meant to be perfectly straight. 
    \item Imperfect shapes - a shape is intended to be a rectangle but it is slightly off.
    \item Translation - two shapes which were intended to be aligned with each other are drawn misaligned.
    \item Scaling - two shapes which were intended to be identically sized, are drawn slightly differently sized.
\end{itemize}

Rotation and imperfect shapes are corrected as a side effect of the detection procedures (section \ref{section:element_detection}). For rotation, rather than taking the exact line of the contour we instead take the minimum bounding box around the shape (aligned with the page). This effectively corrects rotation and aligns shapes to the vertical axis. Imperfect shapes are corrected due to the tolerances built into the detection (section \ref{section:element_detection}).



Translation and scaling issues are mitigated by adjusting the position and size of objects using average bin packing.

The final output is the wireframe structure in the form of the DSL (described in section \ref{section:framework}) which can be converted directly to HTML by the framework.

Note that a limitation of this approach is that we assume that the wireframe did not intend for these errors but it may be the case that they were intended. We justify this because the majority of websites do not include these inconsistencies. This based on our experience with website design as well as the 1,750 website templates we viewed during the creation of the dataset.

\section{Approach 2: Deep learning segmentation} \label{section:approach2}
The goal of this approach is to use deep learning to perform equivalent tasks to the previous approach (section \ref{section:approach1}) i.e. to translate a sketched wireframe into code. 

We considered three approaches using deep learning:

\begin{enumerate}[label=(\roman*)]
    \item Using convolutional neural networks to categorise elements and containers. This would involve using CNNs to classify both elements and containers. State of the art networks such as Yolo \cite{yolo} or Faster R-CNN \cite{FasterRCNN} can classify and detect accurate bounding boxes in real time. However, this approach would require additional steps for normalisation and hierarchical tree detection.
    
    \item Using an ANN to learn the relationship between an image of a wireframe and the code directly. This has the benefit of being a complete solution with no post processing steps. Pix2code \cite{pix2code} implemented this approach in a similar domain. Pix2code's approach of using a CNN and \textit{long short-term memory} (LSTM) \cite{hochreiter1997long} could be modified for this problem. However, we identified challenges with this approach:
 
    \begin{itemize}
        \item Pix2codes architecture was limited to only 72 tokens, this was due to a fixed memory length in the LSTM. This is because HTML is heavily nested and in order to predict the closing tag the LSTM would have to have the opening tag in its memory. Modern websites contain 100s of elements, and therefore the memory length would have to be dramatically increased. However, increasing the memory length would require considerably more data.
        \item The cost of misprediction was large in this architecture due to the nature of code having to match strict syntax, a misplaced token could result in invalid code which does not compile.
        \item Although pix2code displayed promising results in its synthetic dataset it did not generalise to real websites. Element positions and widths were not taken into account, rather it predicted positions on Bootstrap's grid system \cite{bootstrap_grid}. We believe that in order to make this application general, a dataset of clean code orders of magnitude larger than pix2codes dataset would be required. The collection of which is out of the scope of this dissertation.
    \end{itemize}
    
    \item Using an ANN to learn the relationship between an image of a wireframe and an image of the result. We would use an ANN to translate an image of a wireframe into an image which represents the structure. The structure image would be equivalent to a normalised website (see section \ref{section:dataset}). The normalised form would classify elements and containers as well as mitigating human errors. We would apply a post-processing step to translate the resulting image into code (much the same way we did in our dataset collection in section \ref{section:dataset}).
\end{enumerate}

We chose (iii) - using an ANN to translate a wireframe into a normalised image - as it offered a complete (detection, classification, and normalisation) and feasible solution. We considered designing our own network but we found that existing segmentation networks could be used for this process. Using existing segmentation networks had the advantage of potentially increased performance as a large body of research already exists designing and optimising these networks.

Segmentation may not an intuitive choice for this problem but it can be used for:

\begin{itemize}
    \item Element detection - a segmentation network groups pixels related to an object together. These pixel boundaries can be extracted as element boundaries.
    \item Element classification - a segmentation network will label related groups. Labels correspond to classes from the training set. As such, the network classifies elements.
    \item Element normalisation - the pixel boundaries do not have to directly match the actual boundaries on the sketch. Alterations such as rotation and scaling can be corrected if the normalised version contains these corrections.
\end{itemize}

Segmentation cannot be directly used to perform structural detection but we make use of the process described in section \ref{section:dataset_tree} to perform structural detection from the output image.

\subsection{Preprocessing}

The dataset (section \ref{section:dataset}) contains sketches and their associated normalised version of the website. In order for the segmentation network to understand the normalised image we converted it into a label map i.e. label each pixel with the element class it represents. As our normalised images are 3 channel RGB images we created a new single channel image and converted each RGB colour into a single value. We use values 0 to 10 to represent each element label, as well as the container labels and the background label.

We also resized both the sketch and normalised image to 256x256 in order to decrease the size of our network and therefore decrease the amount of time it would take to train.

\subsection{Segmentation}

We fine tuned an Xception model \cite{xception} (a CNN based on Inception v3 \cite{szegedy2016rethinking}) pretrained on ImageNet \cite{imagenet} - a large real world image database - to use as a backbone network for Deeplab v3+ \cite{deeplab}. The rational for using a pretrained model and fine tuning was:

\begin{itemize}
    \item We had neither the resources nor a sufficiently large dataset to allow training from scratch.
    \item Utilising an existing model trained on ImageNet will have already learnt basic features such as edges and corners giving the network a head start when learning the more abstract features.
    \item Although ImageNet contains mostly real world full colour images, and our dataset contains black and white 2D sketches the basic features in the first layers of the network have learnt transferable features. Further, we were not aware of any other suitable pretrained model for 2D sketches which was compatible as a backbone of Deeplab v3+.
\end{itemize}

Deeplab had provided values for network hyperparameters which lead to peak performance on the PASCAL VOC 2012 dataset \cite{pascal-voc-2012}. We based our hyperparameters on these but made some key adjustments. Deeplab suggested training with a learning rate of 0.01 and multiplying the learning rate by 0.1 every 2000 iterations (to ensure that the algorithm reaches a global minimum as soon as possible) and using a momentum of 0.9 and weight decay of 0.0005. As we were fine tuning rather then training the network from scratch we used a lower learning rate of 0.001. This is a common practice as we expect the pre-trained weights to be quite good already as compared to randomly initialized weights and we do not want to distort them too quickly and too much. We set atrous rates to 12, 24, and 36  (for atrous spatial pyramid pooling \cite{aspp}) and found this gave the CNN very wide spatial context which aids header and footer container classification accuracy. Finally, we used an output stride of 8 (ratio of input to output spatial resolution). We found that these initial parameters gave us a good training time without compromising on accuracy too much but we recognise that better accuracy may be possible with a much larger batch size and increased spatial resolution.

We trained on 1,250 images from our dataset with mini-batches of 3 images. We used early stopping to prevent overfitting and stopped training when test accuracy began to drop.

\subsection{Post-processing}
\begin{figure}[H]
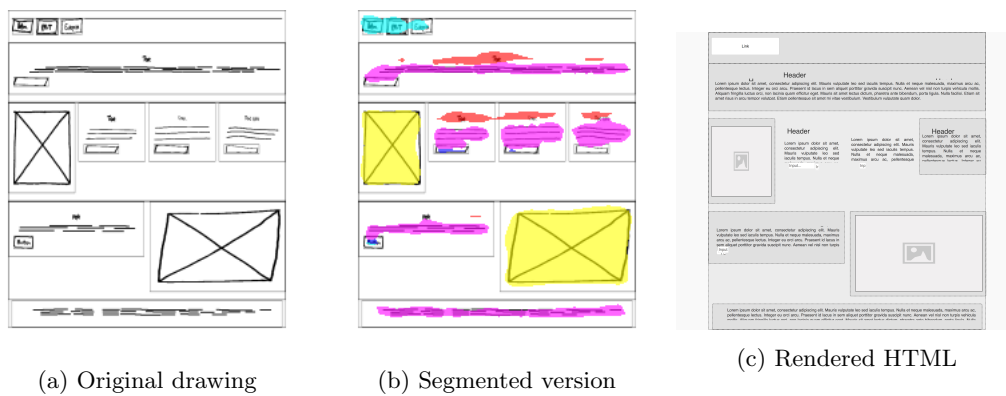

    \begin{subfigure}[H]{0.3\textwidth}
        \includegraphics[width=\textwidth]{assets/method/rendered_page_2_drawn.png}
        \caption{Original drawing}
        \label{figure:approach2_od}
    \end{subfigure}
    \begin{subfigure}[H]{0.3\textwidth}
        \includegraphics[width=\textwidth]{assets/method/rendered_page_2_seg.png}
        \caption{Segmented version}
        \label{figure:approach2_seg}
    \end{subfigure}
    \begin{subfigure}[H]{0.3\textwidth}
        \includegraphics[width=\textwidth]{assets/method/rendered_page_2.png}
        \caption{Rendered HTML}
        \label{figure:approach2_html}
    \end{subfigure}
    \caption{A sample showing how a sketch is segmented and transformed into a website. (a) shows the original wireframe sketch. The sketch is resized to 256x256 and fed into the segmentation network. The network outputs a single channel image containing pixel labels for each element class but we colourise and overlay above the original sketch to visualise see (b). From the network output we apply a post-processing step and use container classification to produce bounding boxes for each element. The tree of elements is fed into the post-processing phase of the framework (section \ref{section:framework}) producing the rendered page (c).}
    \label{fig:deeplearning_sample}
\end{figure}

The result of the segmentation is a single channel image with pixels labeled from 0 to 10 matching the input labels from the pre-processing step. Figure \ref{fig:deeplearning_sample} shows an example of a partially colourised result. 

Elements may contain holes or non perfect edges, as such we filter each element in turn and apply a closing operation to close small gaps, and a erosion operation to remove single pixel lines which connect multiple objects. We then apply contour detection and use the bounding boxes as the elements dimensions. From our list of elements, we apply algorithm 1 to create the hierarchical tree structure. The hierarchical tree DSL is then fed into the post-processing phase of the framework (section \ref{section:framework}) in order to produce the HTML.

\clearpage
\part{Evaluation} \label{section:evaluation}
The goal of our study is to measure performance of each of our approaches in order to identify strengths and weaknesses of each approach in comparison to one another. This comparison will involve two parts: measuring the micro performance of different aspects of each approach, and measuring the macro global performance of each.

The context of this study consists of 250 generated sketched wireframes and their corresponding websites from our dataset (section \ref{section:dataset}) which were withheld from training in both approaches. And 22 hand drawn sketched wireframes drawn by subjects of the trial.

In designing our study we have been influenced by similar research \cite{REMAUI, Moran2018MachineLP, SketchWizard, pix2code, DENIM, SILK} to include particular metrics to aid future comparison. However, we found no other approaches which was directly comparable.

Note that we restrict our evaluation purely to our two approaches and do not include any evaluation of our framework.

\section{Empirical Study Design}

\subsection{Micro Performance}
A disadvantage of some machine learning approaches is the inability to compartmentalise aspects of the system. As such, we are not able to directly compare the normalisation and structural aspects of each approach. However, their performance is taken into account when studying the macro performance of both approaches.

We compare the detection and classification of elements and containers with the aim of identifying strengths and weaknesses of each approach. We aim to answer the following research question:

\textbf{RQ\textsubscript{1}} How well does each method detect and classify elements and containers?

In order to answer RQ\textsubscript{1} we will run each approach over each of the 250 sketched wireframes. As these sketches have been generated from real websites we know the exact x, y, width, height, and label of every element. For each element and container class we will count the number of matches each approach produces compared with the real elements from each sketch.

We define a match if the detected elements x, y, width, height and label equals the actual elements x, y, width, height, and label. A 5\% margin of error on each dimension is allowed, this is because during the sketching process elements were scaled by a random value in the range of $\pm2.5\%$ (see section \ref{section:background}).

We use sketches rather than images of individual components as we want to evaluate the performance in the expected environment i.e. potentially noisy sketches with many other similar looking elements ranging in sizes and rotation.

We measure the performance of detection and classification together, this is because these operations cannot be separated from each other. In the first approach classification and detection are reliant on each other, and in the second approach due to the use of machine learning they happen in parallel.

We calculate the macro mean precision, recall, and F1 score \cite{goutte2005probabilistic}.

\[ Precision=\frac{TP}{TP+FP} \]

\[ Recall=\frac{TP}{TP+FN} \]

\[ F_1=2 \cdot \frac{precision \cdot recall}{precision + recall} \]

where $TP$ corresponds to \textit{true positives} - instances where the bounding box matches. $FP$ correspond to \textit{false positives} - instances where there is no corresponding true element. $FN$ correspond to \textit{false negatives} - instances which were not detected.

These indicators were used as they offer explanation on which aspects of each approach is performing well or poorly. The precision is an indicator used to measure the exactness of each approach i.e. a high precision indicates a low number of false positives. The recall is a measure of completeness i.e. a low recall indicates many false negatives. While precision and recall help explain each model, we use the F1 score as the single metric of performance. The F1 score is the harmonic mean of recall and precision i.e. it conveys the balance between them, a high F1 score indicates the system is both exact and complete.

Note that while other similar research \cite{MobiDev, Moran2018MachineLP} uses ROC curves and confusion matrices \cite{nasrabadi2007pattern} these are not applicable in our case as we are measuring both detection and classification rather then just classification.

\subsection{Macro Performance}

The goal of measuring the macro performance of each approach is to identify strengths and weaknesses of each approach. We answer two research questions:

\begin{itemize}
    \item \textbf{RQ\textsubscript{2}} How well does the generated website match the structure of the original website based on the sketch?
    \item \textbf{RQ\textsubscript{3}} How well does each approach generalise to unseen examples i.e. examples not synthetically sketched?
\end{itemize}

For RQ\textsubscript{2} we used three methods to measure performance:
\begin{itemize}
    \item Visual comparison
    \item Structural comparison
    \item User study
\end{itemize}

We use three methods for two reasons. Firstly, we have not found a perfect measure of performance - there are advantages and disadvantages of each method (explained below). And secondly, we include metrics which other similar research has included to allow future research to compare against our results.

Where appropriate we conduct a one-tailed Mann-Whitney U test \cite{mann1947} to determine if the results are statistically significantly different. Results are declared as statistically significant at a 0.05 significance level. Mann-Whitney U is a nonparametric test. It assumes observations are independent, responses are ordinal, and the distributions are not normally distributed. We use the following formulation:

\[ U=n_1 n_2 \frac{n_2(n_1 + 1)}{2} - \sum_{i=n_1 + 1}^{n_2} R_i\]

where $U$ = Mann-Whitney U test, $n_1$ = sample size one, $n_2$ = sample size two, and $R_i$ = Rank of the sample size.

We use the result of the Mann-Whitney U test to calculate the \textit{p-value} \cite{greenland2016statistical}. The p-value is the probability under the null hypothesis of obtaining a result equal to or more extreme than what was actually observed. A p-value $\leq$ 0.05 indicates strong evidence against the null hypothesis.

Statistical significance testing tells us there was a difference between two groups. We calculate the effect size to indicate how big the effect was. We use \textit{Cliff’s delta} (d) \cite{cliff1993dominance} which measures how often one value in one distribution is higher than the values in the second distribution

\subsubsection{RQ\textsubscript{2} - Visual comparison} \label{section:evaluation_RQ2_visual}

Following from \cite{REMAUI, Moran2018MachineLP} we use the \textit{structural similarity} (SSIM) \cite{ssim} and \textit{mean squared error} (MSE) \cite{MSE} as metrics to evaluate the pixel level visual similarity between the generated websites and the original websites. We will run each approach over all the 250 sketched wireframes and calculate the SSIM and MSE over the generated and original website.

SSIM is a perception-based model that considers image degradation as perceived change in structural information. MSE measures the mean squared per pixel error between two images, a MSE of 0 is for identical images. An SSIM of 1 indicates exact structural similarity between two images.

As the original website contains stylistic differences (e.g. colours, fonts) while our generated HTML represents purely the structure of the wireframe, we use our normalised version of the original website which normalises all style leaving only the structure of the website.

Pixel level comparison has a number of flaws \cite{MSE_flaws}, most notably is that slight differences in structure - which a human would view as benign - lead to widely skewed results. As such, we do not agree with other research \cite{REMAUI, Moran2018MachineLP} that this is a good measure of performance. Therefore, while we measure the pixel level performance to aid comparison future research, we do not interpret the results.

For both MSE and SSIM we conduct two hypothesis tests,

\[ \textrm{MSE} \qquad H_0: m_1 = m_2, H_1: m_2<m_1 \]
\[ \textrm{SSIM}  \qquad H_0: m_1 = m_2, H_1: m_1<m_2 \]

where $m_1$ is the mean of approach 1 and $m_2$ is the mean of approach 2. These tests will indicate if our deep learning approach outperforms our classical computer vision approach. This test is significant as it indicates if deep learning is a promising potential research direction or not.

\subsubsection{RQ\textsubscript{2} - Structural comparison}

We compare the generated structure with the structure extracted from the normalised image. Note that due to websites having many possible structures which lead to the exact same rendered result we do not directly use the source code from the website (as discussed in section \ref{section:dataset}). Instead we create a normalised version and extract the structure from this. This process results in a more consistent structure i.e. identical rendered web pages have identical structures. As both approaches have either been trained or use the same process to produce the generated structure the results are comparable.

We use the WagnerFischer \cite{WagnerFischer} implementation of Levenshtein edit distance \cite{Levenshtein} to compare the hierarchical tree similarity. We implemented the pre-order traversal to deconstruct trees.

The edit distance is a useful performance metric as:

\begin{itemize}
    \item It indicates how each approach is making mistakes i.e. insertions, deletions, or edits.
    \item Gives an absolute value of performance, i.e. zero edits is a perfect score.
    \item Allows comparison between the two methods
\end{itemize}

However, the edit distance fails to capture the significance of edits i.e. a large element missing is seen as more significant then a smaller element missing to a human observer.

We conduct a hypothesis test on the total number of operations,

\[ H_0: m_1 = m_2, H_1: m_1<m_2 \]

where $m_1$ is the median number of operations of approach 1 and $m_2$ is the median number of operations of approach 2. This test will indicate if our deep learning approach outperforms our classical computer vision approach. This is test is significant as it indicates if deep learning is a promising potential research direction or not.

\subsubsection{RQ\textsubscript{2} - User study} \label{section:RQ2_user_study_design}

\begin{figure}[H]
    \centering
    \includegraphics[width=0.5\textwidth]{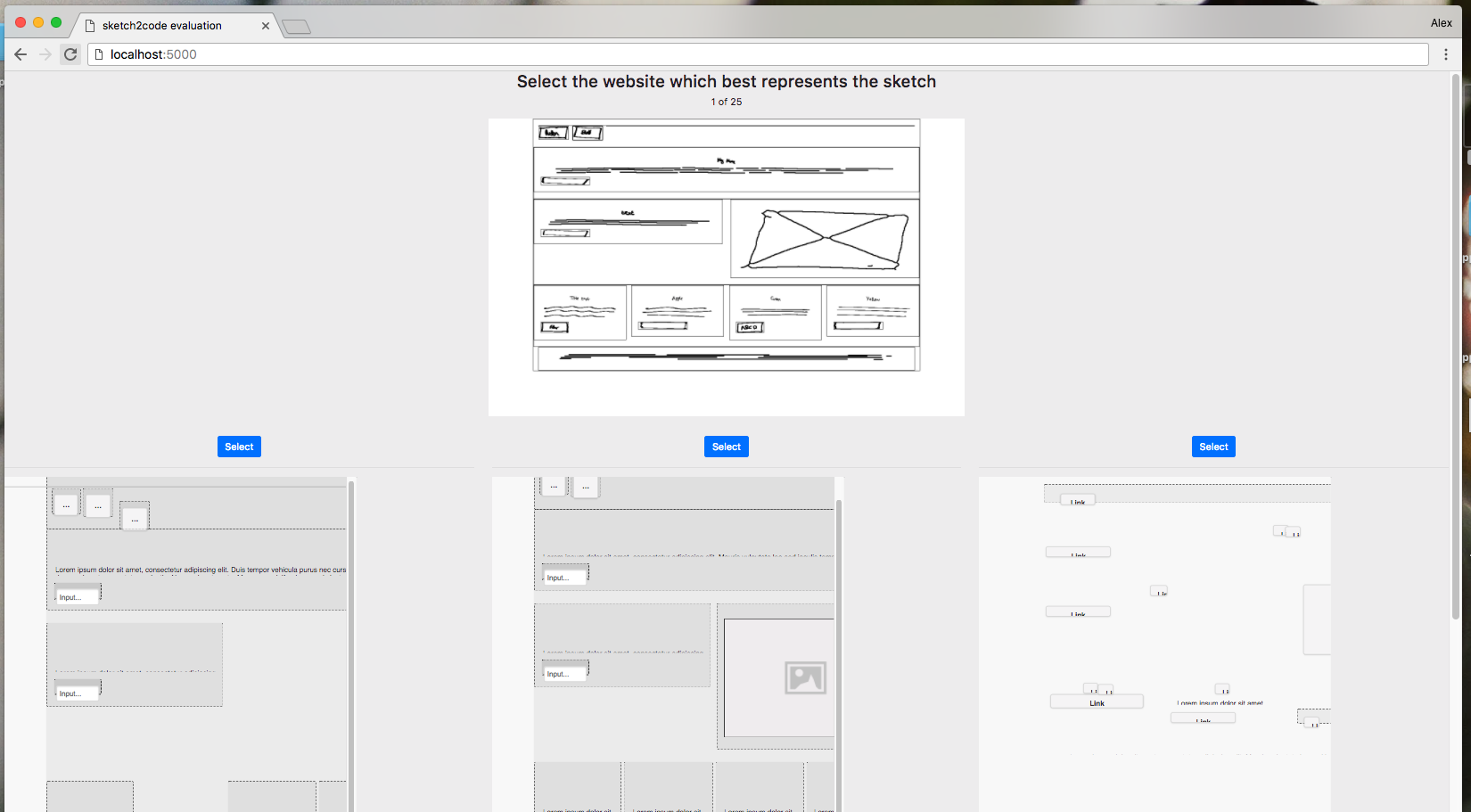}
    \caption{User study evaluation interface. A subject is presented with a wireframe sketch and must select the website which best represents the sketch from the three choices. The choices consist of a generated website by approach 1, a generated website by approach 2, and a control website which was generated from a completely different sketch. The order of the three choices is randomised. Subjects perform the test on 25 randomly selected wireframes from the evaluation set (all subjects evaluate the same websites but order is randomised).}
    \label{fig:user_study_interface}
\end{figure}

Our final method of answering RQ\textsubscript{2} is a small scale user study. Our aim is to understand which approach performs better.

To do this we presented our test subjects with 25 randomly selected sketches from our 250 sketched evaluation set. For each sketch we asked them to select the webpage which best represented the wireframe sketch. The three choices consist of a webpage generated by approach 1, a webpage generated by approach 2, and a control of a randomly generated webpage from another sketch. Figure \ref{fig:user_study_interface} shows the user interface.

Our population consists of 22 website experts - 9 website designers and 13 developers - from Bird Marketing \cite{BirdMarketing} - a digital marketing and website development company. Website experts were chosen as the population as they would be best able critically evaluate the websites.

The study was double blinded with choices and the order of sketches being randomised per subject. Subjects were instructed not to talk and performed the study in isolation. The first three sketches at the beginning of the test were used for calibration of the subjects and discarded from the results.

From the results we calculated the preference (choice which received the highest votes) and preference strength per subject i.e. the difference between the top preference (by total number of votes) and the sum of the other choices over the number of sketches sampled. The preference strength metric yields $0\leq s \leq 1$ which indicates how strong a subjects preferences towards one choice over the others was i.e. 0 indicates no preference and 1 indicates total preference.

\subsubsection{RQ\textsubscript{3} - User study}

The aim is to evaluate how well each method generalises to unseen examples. The purpose of this is to identify if any bias exists in our sketching process.

To do this we conduct a qualitative user study whereby we ask subjects to draw a website wireframe. Subjects are only instructed to use the predefined symbols described in section \ref{section:dataset}. Subjects use the same materials, A4 white paper with black ball point pen for the wireframe. Wireframes were collected, photograph, and fed into both approach 1 and approach 2. We used the same method as described in the RQ\textsubscript{2} user study - displaying three choices to each subject and asking for their preference.

The study was double blinded with choices and the order of sketches being randomised per subject. Subjects were isolated and instructed not to talk while performing the study (both drawing and scoring). We included an additional three sketches at the beginning of the test for calibration, the results of which were discarded from the results. A subjects own sketch was also excluded from the evaluation.

We used the same population described in the RQ\textsubscript{2} user study - 22 website experts. This introduces a non-independent population as they had already participated in the previous study. However, we do not see this as a significant issue as the subjects participation in the previous study does not influence their ability to evaluate which generated website best represents the wireframe.

\section{Study Results}

\subsection{Micro Performance}

\begin{table}[h!]
    \centering
    \begin{tabular}{ |c|c|c|c| }
     \hline
     \textbf{Component} & \textbf{Precision} & \textbf{Recall} & \textbf{F1 score} \\
     \hline
     Image & 1.000 & 0.351 & 0.520\\
     \hline
     Input & 0.190 & 0.556 & 0.283 \\
     \hline
     Button & 0.424 & 0.419 & 0.421 \\
     \hline
     Paragraph & 0.747 & 0.461 & 0.570 \\
     \hline
     Title & 0.651 & 0.447 & 0.530 \\
     \hline
    \end{tabular}
    \caption{Approach 1 element classification and detection results (250 samples). Two highlighted results are images having a precision of 1. This shows that images had zero false positives. However, images also had a low recall i.e. lots of false negatives. This indicates that approach 1's image classifier was very highly discriminatory. Another highlighted result is the input class having a precision of 0.190 i.e. many false positives. We suggest that this is due to container elements being misclassified as input elements indicating the need for better post detection filtering.}
\label{table:RQ1_approach1_element_results}
\end{table}

\begin{table}[h!]
    \centering
    \begin{tabular}{ |c|c|c|c| }
     \hline
     \textbf{Component} & \textbf{Precision} & \textbf{Recall} & \textbf{F1 score} \\
     \hline
     Image & 0.896 & 0.741 & 0.811\\
     \hline
     Input & 0.712 & 0.601 & 0.652 \\
     \hline
     Button & 0.772 & 0.597 & 0.673 \\
     \hline
     Paragraph & 0.562 & 0.461 & 0.548 \\
     \hline
     Title & 0.627 & 0.734 & 0.676 \\
     \hline
    \end{tabular}
    \caption{Approach 2 element classification and detection results (250 samples). Generally these outperform approach 1's results with the exception of the paragraph class. The precision score of the paragraph element performed worse in approach 2 compared with approach 1 (0.562 vs. 0.747). This indicates that approach 1's paragraph detection produced fewer false positives.}
\label{table:RQ1_approach2_element_results}
\end{table}

Table \ref{table:RQ1_approach1_element_results} and \ref{table:RQ1_approach2_element_results} show the precision, recall, and F1 scores of both approaches. Approach 2 achieved a significantly higher F1 score on classification of all elements except paragraphs. This indicates that our implementation of deep learning segmentation outperforms classical computer vision techniques at the task of element detection and classification. This result is unsurprising as deep learning techniques have shown to outperform classical techniques in detection and classification problems \cite{krizhevsky2012imagenet,szegedy2015going,DBLP:journals/corr/ZeilerF13,DBLP:journals/corr/SimonyanZ14a}.

\begin{figure}[H]
    \centering
    \begin{subfigure}[H]{0.49\textwidth}
        \includegraphics[width=\textwidth]{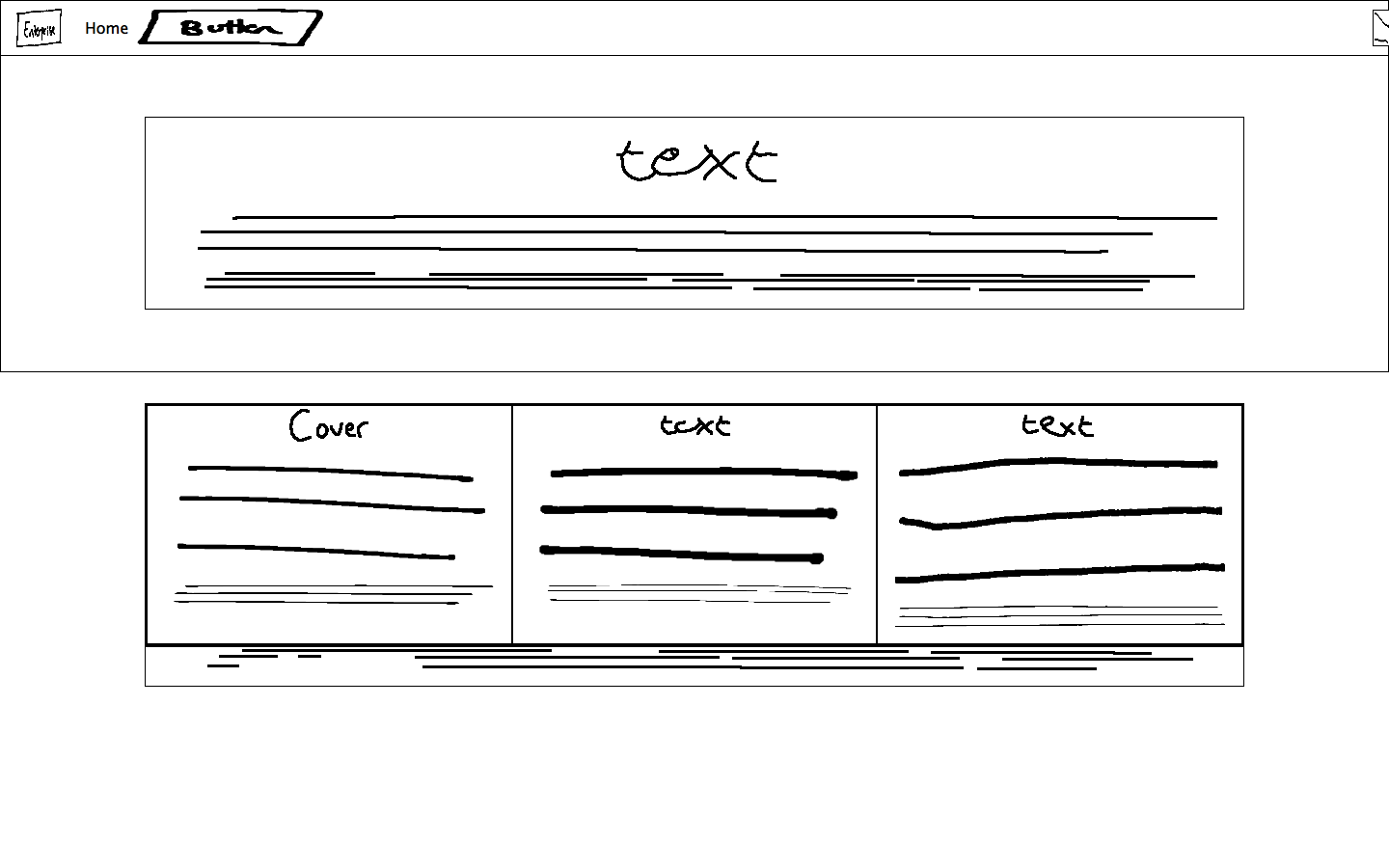}
        \caption{Sketched version}
    \end{subfigure}
    \begin{subfigure}[H]{0.49\textwidth}
        \includegraphics[width=\textwidth]{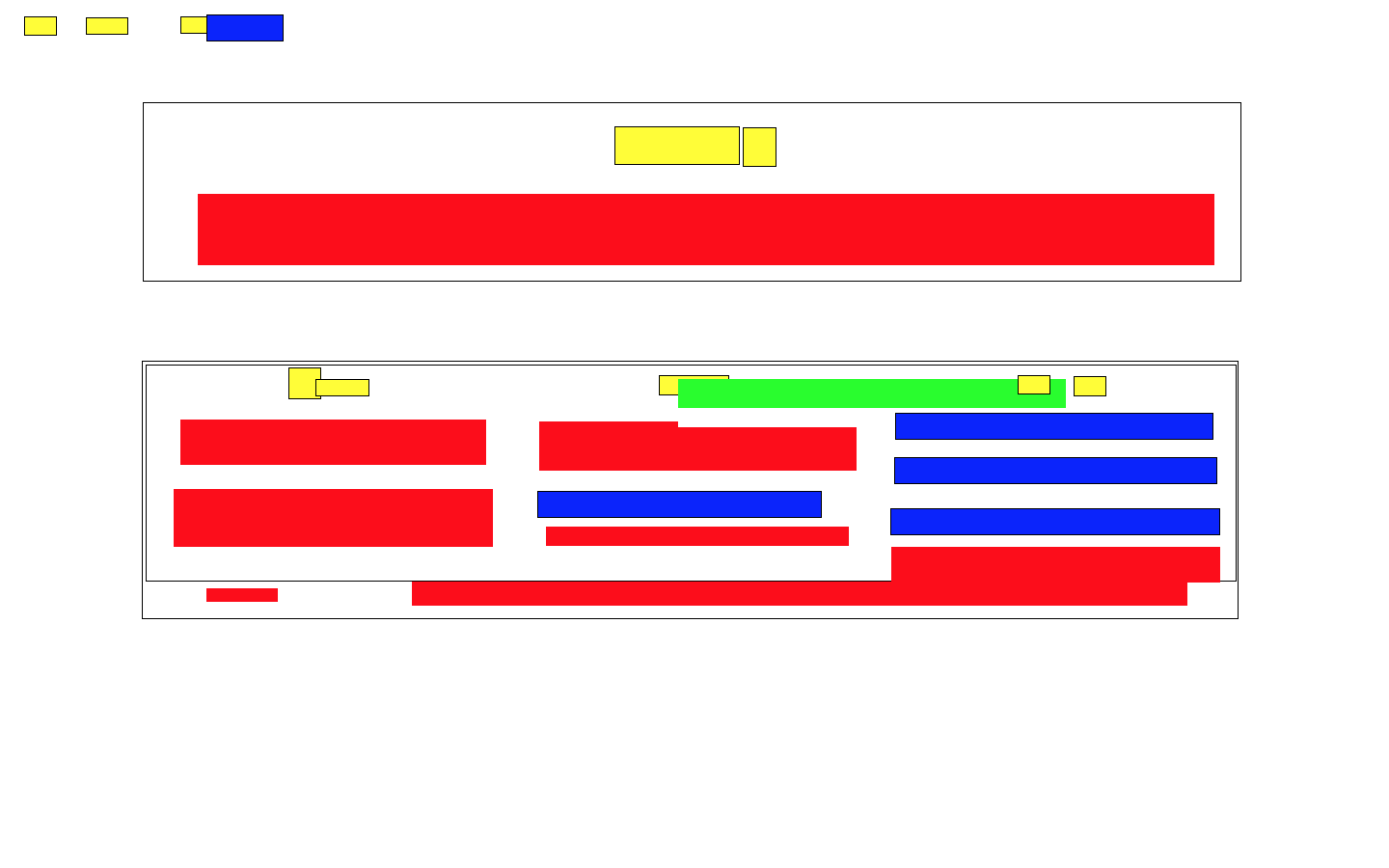}
        \caption{Generated normalised version}
    \end{subfigure}
    \caption{One explanation of approach 1's poor performance metrics is that it performs poorly when elements are close together. These images show a particularly poor case. The three boxes along the bottom of the sketch (left image) confuse the classifier resulting in merged and distorted result (right image).}
    \label{fig:rq1_e1_close_together}
\end{figure}

\begin{figure}[H]
    \centering
    \begin{subfigure}[H]{0.25\textwidth}
        \includegraphics[width=\textwidth]{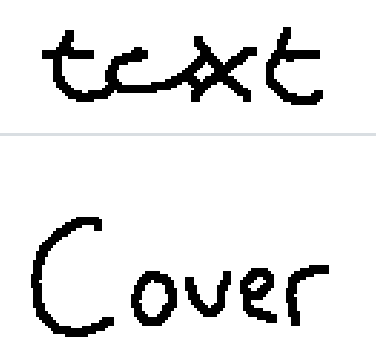}
        \caption{Drawn Text}
    \end{subfigure}
    \begin{subfigure}[H]{0.25\textwidth}
        \includegraphics[width=\textwidth]{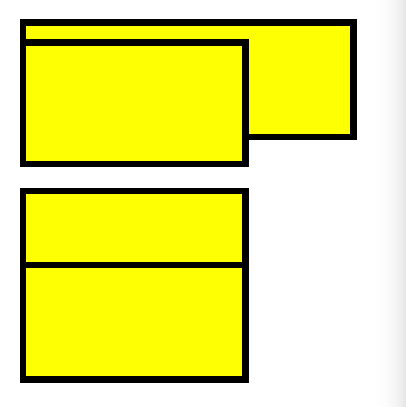}
        \caption{Generated normalised result}
    \end{subfigure}
    \caption{Approach 1 text classifier often performs poor text detection resulting in multiple text boxes. This is another potential cause of approach 1's poor performance.}
    \label{fig:rq1_misclassification}
\end{figure}

Table \ref{table:RQ1_approach1_element_results} shows the results of approach 1. Below we hypothesis features of approach 1 which aid explanation of these results.

\begin{itemize}
    \item Poor precision of the input element class may be due to the similarity between input elements and containers. This score indicates better post filtering is required.
    \item The image element class has low recall. This indicates that the image classifier is too discriminatory and that it may be able to achieve better performance by relaxing its detection constraints.
    \item The classifier was sensitive to element size. Both overly small and large elements were misdetected as multiple elements which reduced performance across all element classes. This appears to be especially prevalent in the paragraph element class. When the element is large it is detected as multiple input elements rather then a single paragraph element.
    \item Element close together would often be merged as seen in figure \ref{fig:rq1_e1_close_together}.
    \item The text detector added additional elements. Figure \ref{fig:rq1_misclassification} shows a sample from the validation set. The SWT text detector often failed to merge close text boxes, this resulted in multiple detections for a single element containing text.
\end{itemize}

\begin{figure}[H]
    \centering
    \begin{subfigure}[H]{0.49\textwidth}
        \includegraphics[width=\textwidth]{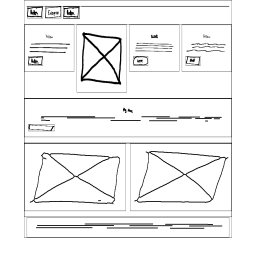}
        \caption{Sketched version}
    \end{subfigure}
    \begin{subfigure}[H]{0.49\textwidth}
        \includegraphics[width=\textwidth]{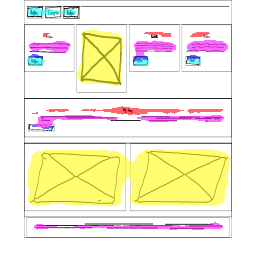}
        \caption{Generated normalised version}
    \end{subfigure}
    \caption{Demonstrates problems which arose in approach 2's classification. While element classes were often correct, the bounding boxes were often slightly off. We hypothesise that this is a primary reason for a non-optimal performance in approach 2. (b) demonstrates how close together element classes merged i.e. the two images are joined (yellow). Further, due to our normalisation of title elements (red) we extend the title element to 100\% width. This appears to result in patchy detection which has a negative impact performance by increasing false positives.}
    \label{fig:rq1_e2_joined_classes}
\end{figure}

Approach 2 performed significantly better compared to approach 1. Approach 2 did not suffer the same problems as approach 1 e.g. approach 2 did not suffer from a low input element class precision score. However, approach 2 did not perform optimally ($F_1$ score of 1 across all element classes). We have identified a number of issues the classifier faced which may explain non-optimal performance:

\begin{itemize}
    \item Merged classes. The classifier would merge elements which were close together. This can be seen in figure \ref{fig:rq1_e2_joined_classes}. Our post-processing procedure attempted to mitigate this issue by eroding the image after classification. However, in some cases this procedure was not enough and resulted in misdetections.
    \item Button and input confusion for small elements. We identified that small button elements would be misclassified as input elements. We hypothesise that this is due to the text inside of the buttons losing 'text like' characteristics at very small scales resulting in misclassification.
\end{itemize}

\begin{table}[h!]
    \centering
    \begin{tabular}{ |c|c|c|c| }
     \hline
     \textbf{Component} & \textbf{Precision} & \textbf{Recall} & \textbf{F1 score} \\
     \hline
     Header & 1.000 & 0.888 & 0.941 \\
     \hline
     Row & 0.674 & 0.738 & 0.705 \\
     \hline
     Stack & 0.705 & 0.779 & 0.740 \\
     \hline
     Footer & 1.000 & 0.950 & 0.974 \\
     \hline
     Form & 0.898 & 0.344 & 0.497 \\
     \hline
    \end{tabular}
    \caption{Approach 1 container classification and detection results (250 samples). Header and footer elements had the highest performance. This may be explained by their distinct nature (100\% width at top or bottom of page). Form elements had the lowest performance due to poor recall (0.344). This indicates that the classifier was highly discriminatory.}
\label{table:RQ1_approach1_container_results}
\end{table}

\begin{table}[h!]
    \centering
    \begin{tabular}{ |c|c|c|c| }
     \hline
     \textbf{Component} & \textbf{Precision} & \textbf{Recall} & \textbf{F1 score} \\
     \hline
     Header & 0.701 & 0.854 & 0.770\\
     \hline
     Row & 0.591 & 0.485 & 0.533 \\
     \hline
     Stack & 0.425 & 0.523 & 0.469 \\
     \hline
     Footer & 0.800 & 0.831 & 0.815 \\
     \hline
     Form & 0.683 & 0.452 & 0.544 \\
     \hline
    \end{tabular}
    \caption{Approach 2 container classification and detection results (250 samples).}
\label{table:RQ1_approach2_container_results}
\end{table}

The results shown in table \ref{table:RQ1_approach1_container_results} and \ref{table:RQ1_approach2_container_results} show that approach 1 performed significantly better at container classification for all elements except forms which it performed marginally worse in. This result is unexpected as we expected the deep learning approach to outperform the classical computer vision approach as it did in element classification. This discrepancy may be explained by the approaches learning from different data. Approach 1 used structural data about the size and position of the container as well as the types of elements inside. Approach 2 used raw pixel data. We hypothesise that as containers are classified by their context (e.g. position on page) rather then visual appearance, approach 2 struggled more using pixel data.

We identified that both approaches routinely misclassified stacks as rows and vice versa. This may explain the lower performance for these elements. This indicates that the machine learning in both methods did not learn the distinctive difference between these two elements i.e. rows contain elements horizontally while stacks contain elements vertically. We hypothesise the better feature engineering may alleviate this issue.

\fbox{\begin{minipage}{\textwidth}
\textbf{RQ\textsubscript{1}}: Approach 2 outperformed approach 1 in element classification and detection while approach 1 outperformed approach 2 in container classification.
\end{minipage}}

\subsection{Macro Performance}

\subsubsection{RQ\textsubscript{2} Visual comparison}

\begin{figure}[H]
    \centering
    \includegraphics[width=0.7\textwidth]{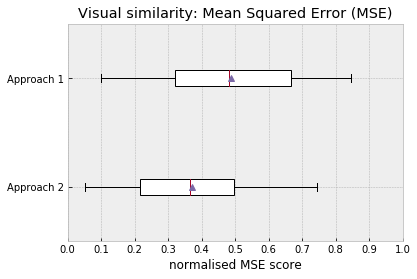}
    \caption{Mean squared error of approach 1 and approach 2 (250 samples). The purple triangle represents the mean while the red line represents the median. Approach 2 has a lower median then approach 1 indicating better performance.}
    \label{fig:RQ2_visual_MSE}
\end{figure}

\begin{figure}[H]
    \centering
    \includegraphics[width=0.7\textwidth]{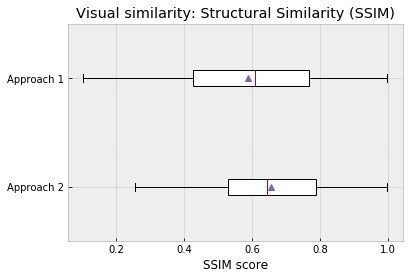}
    \caption{Structural similarity of approach 1 and approach 2 (250 samples). The purple triangle represents the mean while the red line represents the median. Approach 2 has a lower higher median then approach 1 indicating better performance.}
    \label{fig:RQ2_visual_SSIM}
\end{figure}

For MSE $p < 0.001$ indicating strong evidence to reject $H_0: m_1 = m_2$ favoring the alternative $H_1: m_2 < m_1$. We find a relatively strong effect size $d=0.319$.

For SSIM $p = 0.001$ indicating strong evidence to reject $H_0: m_1 = m_2$ favoring the alternative $H_1: m_1 < m_2$. We find a moderate effect size $d=0.156$.

These results indicate that on visual similarity the deep learning approach outperforms classical computer vision approach.

Note that as discussed in section \ref{section:evaluation_RQ2_visual} we have included these metrics to allow future studies to compare results but we do not analyse these results in this dissertation as we do not agree that they are valid tool for comparison.

\fbox{\begin{minipage}{\textwidth}
\textbf{RQ\textsubscript{2}} Visual: Approach 2 outperformed approach 1 in both MSE and SSIM scores.
\end{minipage}}

\subsubsection{RQ\textsubscript{2} Structural comparison}

Note that in all box plots the purple triangle represents the mean, the red line represents the median, and circles represent outliers (outside quartiles).

\begin{figure}[H]
    \centering
    \includegraphics[width=0.7\textwidth]{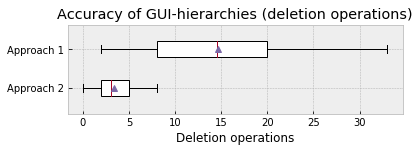}
    \caption{Number of deletion operations (250 samples). Approach 2 has a smaller median and range indicating fewer deletion operations per sketch.}
    \label{fig:RQ2_structural_dels}
\end{figure}

The number of deletion operations for each approach is shown in figure \ref{fig:RQ2_structural_dels}. Deletion operations are required to remove stray elements added. Approach 2 has a smaller median and range, this indicates that this approach does not add many additional elements. On the other hand, approach 1 has a significantly higher number of deletion operations. This is expected based on the RQ\textsubscript{1} results which suggest that approach 1 adds many false positives leading to poor performance.

\begin{figure}[H]
    \centering
    \includegraphics[width=0.7\textwidth]{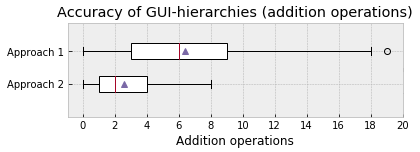}
    \caption{Number of addition operations (250 samples). Approach 2 has a smaller median and range indicating fewer addition operations per sketch.}
    \label{fig:RQ2_structural_adds}
\end{figure}

The number of addition operations for each approach is shown in figure \ref{fig:RQ2_structural_adds}. Addition operations are required to add elements which were not detected. Again, approach 2 appears to outperform approach 1 with a lower median and range. This suggests that approach 2 correctly detects more elements.

\begin{figure}[H]
    \centering
    \includegraphics[width=0.7\textwidth]{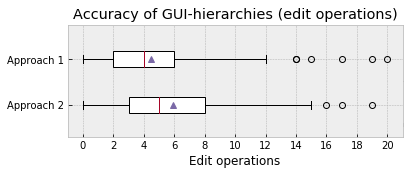}
    \caption{Number of edit operations (250 samples). Approach 1 has a smaller median and range indicating fewer edit operations per sketch.}
    \label{fig:RQ2_structural_edits}
\end{figure}

The number of edit operations for each approach is shown in figure \ref{fig:RQ2_structural_edits}. Edit operations are required to change the label of an element which is placed correctly but has the wrong element class. We see that approach 1 has a lower median and range, this appears to counter the trend shown by the addition and deletion operations. However, as edit operations require the element to be correctly placed but with the wrong label this relationship only indicates that approach 2 had more correctly detected however misclassified elements.

\begin{figure}[H]
    \centering
    \includegraphics[width=0.7\textwidth]{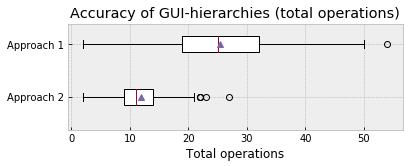}
    \caption{Total number of operations (250 samples). Approach 2 has a smaller median and range indicating fewer operations per sketch.}
    \label{fig:RQ2_structural_total}
\end{figure}

The total number of operations (edits, additions, and deletions) per approach is shown in figure \ref{fig:RQ2_structural_total}. We use total number of operations as a performance indicator, a score of 0 is a perfect replica of a wireframe. Our hypothesis test results in $p<0.001$ indicating strong evidence to reject $H_0: m_1 = m_2$ favoring the alternative $H_1: m_2 < m_1$. We find a large effect size with $d=0.818$. These results indicate that our deep learning approach strongly outperformed the classical computer vision approach. Further, the deep learning approach has a smaller spread which suggests that it performs more consistently.

\fbox{\begin{minipage}{\textwidth}
\textbf{RQ\textsubscript{2}} Structural: Approach 2 outperformed approach 1 with a lower median and more consistent performance.
\end{minipage}}

\subsubsection{RQ\textsubscript{2} User study}

The top preference was approach 2 with 22/22 votes. These results show that the deep learning approach was considered the best performing of the three approaches by all subjects. The mean user preference was 0.76. This shows that 76\% of the time a subject would vote for approach 2 over approach 1 or the control. This indicate that approach 2 strongly outperforms approach 1.

\fbox{\begin{minipage}{\textwidth}
\textbf{RQ\textsubscript{2}} User study: Approach 2 outperformed approach 1 with all subjects preferring it with a strong preference.
\end{minipage}}

\subsubsection{RQ\textsubscript{3} User study}

The top preference was approach 1 with 15/22 votes and then approach 2 with 7 votes. The mean user preference was 0.53. These results show that approach 1 was considered the better approach by most subjects but there was a significantly weaker preference strength compared with the RQ\textsubscript{2} user study. As approach 2 was top performing in the RQ\textsubscript{2} user study these results indicate that approach 2 has poor generalisation to unseen examples.

It is unsurprising that approach 1 generalises better than approach 2 as approach 2 learnt directly from the dataset, and would more easily reflect any biases in the data. While approach 1 was partially a manual process and therefore less likely to reflect biases from within the dataset. We hypothesis that this was due to the small number of drawn elements (90) used to sketch the dataset. A larger number of drawn elements from a wider number of people (to increase the variation in styles) could mitigate this issue. We leave the implementation of this to future research.

Further, the control group got no votes from any subject. This indicates two things: that both approaches do have a positive impacting on sketching (as expected), and that the synthetic sketches are comparable to some degree to real sketches.

\fbox{\begin{minipage}{\textwidth}
\textbf{RQ\textsubscript{3}}: Approach 1 mildly outperformed approach 2 with the majority of subjects preferring approach 1 over approach 2.
\end{minipage}}

\subsection{Discussion}

The results from RQ\textsubscript{1} indicate that the deep learning approach outperforms the classical computer vision approach in element classification but not in container classification. We hypothesised that this is due to elements being distinguished visually which plays to the strengths of CNNs while containers are best distinguished through context information. One possible solution is to perform container classification through a separate network.

The results from all three methods in RQ\textsubscript{2} indicate that the deep learning approach outperforms the classical computer vision approach in macro performance. The results are very promising and provide a strong indication that future research into deep learning approaches in this problem domain may provide solutions with high enough performance to be used in consumer applications.

RQ\textsubscript{3} showed approach 1 outperformed approach 2 on real sketches. This suggests that the dataset sketching procedure requires a greater variety in element sketches to allow the deep learning approach to better generalise to unseen sketching styles.

\clearpage
\part{Conclusion}  \label{section:conclusion_part}
\section{Limitations \& Threats to Validity}

\subsection{Threats to internal validity}

In this section we describe threats to our results validity and explain the steps we took to mitigate these.

\textbf{Dataset}

The normalisation preprocessing step (section \ref{section:dataset}) removes styles and transforms elements on web pages into a consistent size. While developing this process we identified numerous special cases which we added exceptions for e.g. iframes. Failure of the normalisation step due to an unidentified special case may result in an erroneous data entry. Part of the dataset is used for validation and therefore erroneous data should be avoided. To ensure the validity of our dataset, we sampled a random statistically significant segment of our dataset and manually inspected the normalised images. We did not observe any irregularities, thus mitigating a threat related to quality of our dataset.

\lstinline{<header>} and \lstinline{<footer>} HTML elements are not enforced by the HTML compiler and often forgotten. This may degrade the quality of our dataset as we assume well labeled code. In order to mitigate this risk we chose to exclusively use Bootstrap website templates in our dataset. Bootstrap enforces best practices, thus decreasing the risk of mislabeled containers. Again, we sampled a random statistically significant segment of our dataset and manually inspected the website structure. We observed a small number of irregularities and manually corrected the dataset prior to training and validating our approaches.


\textbf{Evaluation methods}

In comparing the macro performance of our two approaches we found no single method which fairly evaluated. As such, we used three methods of comparison.

Rather then directly comparing the generated structure with the pages HTML our structural comparison method compared the generated structure with the extracted structure from the normalised web pages (see section \ref{section:dataset_tree}). Adding this extra step risks systematic error in the evaluation. However the extra step was necessary in order to ensure consistent structure for comparison as there are multiple valid HTML structures which produce the exact same visual result. We mitigate this risk by taking a random statistically significant segment of the dataset and reproduced an image of the normalised website from the extracted structure and manually compared these images to the actual normalised website. We observed no irregularities.

For our user study we used a population of 22 website experts from Bird Marketing. As the entire population came from a single company there is risk selection bias. Bias may come from the company having particular design practices which are not widely used. We believe this does not compromise the validity of our results as we conducted a survey and found that all subjects had worked for at least one other company in a design/developer role. The range of companies subjects had worked for was diverse and so expect design practices from a range of backgrounds.

\subsection{Threats to external validity}

Threats to external validity concern the generalisation of results.
We have implemented two approaches focusing on websites, we assert that implementing our approaches for other application domains (e.g. Android or Desktop) constitutes mainly of an engineering effort.

Our dataset consists of 1,750 website templates which use Bootstrap. Restricting our dataset collection to Bootstrap helped data consistency (see section \ref{section:dataset}) but may introduce a bias into our dataset. We refute this claim as Bootstrap is a structural framework and it does not place any restrictions on the layout of a website, any website can be recreated using Bootstrap \cite{Bootstrap}. 

\subsection{Limitations}

\textbf{Dataset}



Our dataset sketching process (see section \ref{section:dataset}), aims to take a normalised website as input and produce a sketched wireframe. We base our analysis that these sketched wireframes are comparable to wireframes created by humans. While some examples are comparable, others are less comparable. Humans may emphasize different aspects of the wireframe e.g. not put detail into small items compared with large items.


\textbf{Approach 2}

A limitation of this approach is that it is difficult to analysis the micro performance of the detection, classification, and normalisation aspects of this approach. This is due to the black box nature of some deep learning models. Analysing the performance is useful to understand where the network needs to improve and how the network responds to different types of data in order to develop higher performing networks. Methods such as maximal activation \cite{DBLP:journals/corr/GirshickDDM13}, image occlusion \cite{DBLP:journals/corr/ZeilerF13}, and saliency maps \cite{DBLP:journals/corr/SimonyanVZ13} could be used to gain an understanding of which features the network uses to discriminate. However, in our evaluation we chose to focus on comparative techniques. 

\textbf{Evaluation}

Our RQ\textsubscript{3} user study restricted subjects to using element symbols from the list outlined in section \ref{section:background}. Some designers use variants of these symbols, we restricted to this list of symbols as approach 1 was developed to only recognise these symbols and approach 2 was trained from these symbols. This is not a major limitation of our design as both approaches can be modified to add additional symbols. Further, our evaluation framework can easily add and assess new symbols by adding them to the corpus used by the sketching process. 


\section{Conclusion} \label{section:conclusion}

The goal of this thesis was twofold: Create an application which translates a sketched wireframe into a website, and to explore how deep learning compares to classical computer vision methods for this task.

This dissertation has built upon existing research and extended this research to a novel domain of wireframe to code translation. We have presented an end to end framework which translates wireframes into websites and produces results in real time. Section \ref{section:framework} explains how our framework has been developed to be easy to use: by allowing images from web cameras or phone camera; by hosting the rendered website for collaboration; by using common wireframe symbols therefore limiting any training required. We have released a dataset along with tools to recreate or add to the dataset. We have developed two approaches: a classical computer vision approach, and an approach based on deep semantic segmentation networks. Our deep learning approach uses an innovative technique by training on synthetic sketches of website wireframes. Finally, we designed repeatable empirical techniques to evaluate how well a system translates a sketched wireframe into a website. As such, we consider we have achieved our two goals.

Our evaluation shows that neither approach we developed had high enough performance to be used in production environments. However, we assert that our dataset, framework, and evaluation techniques will make a significant impact in the field of design to code techniques. In particular, prior to our work we were not aware of any existing techniques to perform empirical evaluation on translating sketches into code - our dataset sketching procedure allows this. Further, we were not aware of any application of deep learning to this problem domain. We hope this dissertation as well as the release of our dataset and framework will stimulate further research into this domain.

\textbf{Contribution summary:}
\begin{itemize}
    \item The release of out dataset and tools to generate our dataset. We hope this will aid and stimulate future research.
    \item A framework to preprocess images taken from cameras into clean sketches which can be feed into our two applications. As well as translating the DSL produced by these applications into HTML and deploying it live to multiple clients.
    \item An application which uses a classical computer vision approach to translate wireframes to code. This approach achieved moderate performance.
    \item An application which uses a novel deep learning approach to translate wireframes to code. This approach outperformed the classical approach in some domains and is a promising direction for future research.
    \item The design of empirical techniques to analyses the micro and macro performance of wireframes to code applications.
\end{itemize}

\section{Future work}

Our evaluation shows that neither approach we developed had high enough performance to be used in production environments. In this section we discuss directions for future research to increase performance based on what we have learnt from designing and developing our approaches.

During development of approach 1 we identified that matching with pre existing website templates rather than generating a website from scratch would be a direction for future research as it reduces the need for normalisation. Approach 1’s normalisation phase could be replaced by a phase where containers and elements are matched with a library of component templates e.g. an image carousel. Matching with existing templates has the advantage of introducing properties which can be inferred but not directly sketched on paper e.g. animations.

Approach 2 showed significantly higher performance in element detection \& classification with deep learning over approach 1. As we identified in section \ref{section:evaluation} designing the network to perform detection, classification, and normalisation may have decreased its performance. As such, by dividing these problems and using separate networks the performance in each of these tasks may increase. One design for this could be using a CNN such as Yolo \cite{yolo} to perform detection \& classification, and using another network to perform normalisation.

Our approaches focus on single page websites. However, they could be modified to input multi page websites. This would require our framework to be modified to accept multiple images along with information on how to link the pages together. This would allow full websites to be designed using our approaches.

Finally, we focus on websites but we assert that it would primarily be an engineering challenge to use our approach described in other domains such as mobile apps or desktop applications.

\clearpage

\printbibliography

\end{document}